\documentclass[final]{cvpr}

\usepackage{times}
\usepackage{epsfig}
\usepackage{graphicx}
\usepackage{amsmath}
\usepackage{amssymb}
\usepackage[ruled,vlined]{algorithm2e}
\usepackage[dvipsnames]{xcolor}

\newcommand{\tabincell}[2]{\begin{tabular}{@{}#1@{}}#2\end{tabular}}

\usepackage{multicol}
\usepackage{multirow}
\usepackage{color}
\usepackage{comment}
\usepackage{booktabs}
\usepackage{CJKutf8}
\usepackage{threeparttable}
\usepackage{makecell}
\usepackage{tabularx}
\usepackage[pagebackref=true,breaklinks=true,colorlinks,bookmarks=false]{hyperref}

\definecolor{ben}{rgb}{0.9,0.,0.5}

\begin{document}

\title{A Study on Training and Developing Large Language Models for Behavior Tree Generation}

\author{Fu Li$^{1,2}$,  Xueying Wang$^{1,2}$, Bin Li$^{1,2}$, Yunlong Wu$^{1,2, }$\thanks{Corresponding author: yunlong.wu@vip.163.com}, Yanzhen Wang$^{1,2}$, and Xiaodong Yi$^{1}$\\
\\
\normalsize
\textit{$^{1}$Intelligent Game and Decision Lab (IGDL), Beijing 100091, China} \\
\normalsize
\textit{$^{2}$Tianjin Artificial Intelligence Innovation Center (TAIIC), Tianjin 300457, China} \\
}

\maketitle

%%%%%%%%% ABSTRACT
\begin{abstract}\noindent
This paper presents an innovative exploration of the application potential of large language models (LLM) in addressing the challenging task of automatically generating behavior trees (BTs) for complex tasks. The conventional manual BT generation method is inefficient and heavily reliant on domain expertise. On the other hand, existing automatic BT generation technologies encounter bottlenecks related to task complexity, model adaptability, and reliability. In order to overcome these challenges, we propose a novel methodology that leverages the robust representation and reasoning abilities of LLMs. The core contribution of this paper lies in the design of a BT generation framework based on LLM, which encompasses the entire process, from data synthesis and model training to application developing and data verification. Synthetic data is introduced to train the BT generation model (BTGen model), enhancing its understanding and adaptability to various complex tasks, thereby significantly improving its overall performance. In order to ensure the effectiveness and executability of the generated BTs, we emphasize the importance of data verification and introduce a multilevel verification strategy. Additionally, we explore a range of agent design and development schemes with LLM as the central element. We hope that the work in this paper may provide a reference for the researchers who are interested in BT generation based on LLMs.
\end{abstract}
%%%%%%%%% MAIN
\section{Introduction}
Behavior tree is a typical control architecture (CA) that is widely used in computer games and robotics to describe the behaviors of AI and robots \cite{Wu2021}. It can flexibly control task switching between different autonomic agents, such as robots or virtual entities in computer games, and has explicit support for task hierarchy, action sequencing, and reactivity \cite{styrud2022combining}. BT provides a structured and systematic approach to model the behaviors of agents and bridges the gap between abstract goals and concrete implementation of those goals. One of the key advantages of BT is its flexibility in controlling the actions of agents. It offers a modular and hierarchical representation, where each task or behavior can be encapsulated as a separate node in the tree. This allows for feasible modification and extension of the behaviors of agents without disrupting the overall architecture of the system. Compared with other control architectures, BT has the advantages of modularity, hierarchy, reactivity, readability, and reusability \cite{colledanchise2018behavior}.

The researches on BT have been rapidly growing due to its potential in improving complex decision-making processes and enabling more efficient planning and execution of tasks \cite{biggar2021expressiveness}. Existing research can be categorized into several areas. One area focuses on enhancing the implementation of BT itself, such as improving node efficiency \cite{colledanchise2018improving} and robustness \cite{sprague2022adding}. Another area explores methods for efficient BT generation, including learning-based approaches \cite{banerjee2018autonomous, french2019learning, dey2013ql, wathieu2022re} and planning-based approaches \cite{8206502, paxton2019representing, colledanchise2019towards, zhou2019autonomous}. Additionally, there are studies aiming to expand the application scope of BT in various fields \cite{paxton2017costar, shu2019behavior, haijun2019simulation, yao2015adaptive, francillette2020modeling}.

In this paper, we mainly focus on the problem of BT generation. Currently, the vast majority of BTs applied to the real world are manually designed by human experts, which requires rich experience and domain knowledge. 
However, with the increasing complexity and variety of tasks, designing BTs by hand will be a very complex and time-consuming task, which has led to research on automatic BT generation \cite{banerjee2018autonomous, iovino2022survey, colledanchise2018behavior}. Typical automatic BT generation methods include 
planning-based methods \cite{colledanchise2017synthesis,lan2019autonomous,rovida2017extended,colledanchise2019towards}
and learning-based methods \cite{Chen2023, Ahmad2023, Hafner2019, Zhu2019,Cao2023}. 
However, the above methods still have limitations in understanding tasks, especially the ability to generate BTs according to complex task descriptions is still insufficient.

To meet the challenges faced by traditional BT generation methods, we need a method with strong adaptability, generality, and interpretability. LLMs represented by ChatGPT \cite{OpenAI2023GPT4TR} and LLaMA \cite{touvron2023llama_a} are developing explosively, showing strong complex reasoning ability, intelligent emergence ability, and interpretability. The advancement of LLMs presents a promising avenue for streamlining the generation of practical BTs. By harnessing the abilities of these models, it becomes possible to minimize the knowledge requirements associated with this process. Leveraging the power of LLMs can potentially alleviate the burden of extensive task knowledge and intricate sensor coding details that have traditionally been mandatory in the generation of practical BTs. Research on BT generation based on LLM is still in the early stages. In \cite{Cao2022}, a BT embedded approach was proposed to solve the task-level knowledge representation problem. Based on this, a BT generation method based on LLM was proposed, which realizes BT generation according to target task descriptions and few-shot examples through phase-step prompt design \cite{Cao2023}. In \cite{lykov2023llmbrain}, 8.5k self-instruct BT instructor-following data was generated by text-davinchi-003 and used to fine-tune another LLM (Stanford Alpaca 7B) for BT generation.

In general, training and developing LLMs for BT generation is very important.
However, it is still in its infancy. 
Therefore, in this paper, we aim to examine the key technologies involved in various steps. 
The main contributions are summarized as follows:

\begin{itemize}

   \item To the best of our knowledge, limited works focus on training and developing LLMs for BT generation.
   This paper delves into this area to efficiently obtain high-quality BTs. 
   
   \item We introduce a new BT data generation method using Monte Carlo tree search (MCTS) with state representation and operator libraries, improved with a knowledge base and node library constraints, and optimized with verification feedback for reliable the BT generation.
   
   \item We propose a training pipeline based on the foundation LLM and collected BT data set, including pretraining and supervised fine-tuning training, to enhance the abilities of BTGen model in the specialized BT applications.
   
   \item We propose a BT generation framework, named BTGen Agent, to deploy the trained BTGen model in real-world scenarios, which aims to mitigate common issues such as hallucination, data bias, limited out-of-domain knowledge, and issues with less explainability and transparency.
   It comprises four main modules: memory module, action module, planning module, and profile module. 
   In addition to these modules, it also incorporates a refinement mechanism to ensure usability.

   \item We propose a verification and validation (V\&V) driven pipeline to generate high-quality BTs, which runs through all steps in training and developing the BTGen model.
   It ensures that the functions correctly and meets the specified requirements, as well as provides feedback to improve the quality of the BT generation further.
    
\end{itemize}

The rest of the paper is organized as follows. 
First, Section \ref{sec:method} summarizes the methodology of this paper.
Then, Section \ref{sec:foundation_models} introduces existing foundation models and the required abilities for LLMs.
Section \ref{sec:bt_data_collect} shows the basic information of collected BTs for training.
Section \ref{sec:training} and Section \ref{sec:develop} show the pipelines for training BTGen model and developing it, respectively.
Section \ref{sec:vv} analyzes the V\&V on BTGen model and the generated BTs.
Last, Section \ref{sec:open_questions} summarizes the open questions.
Section \ref{sec:conclusion} makes conclusions. 

\section{Methodology}\label{sec:method}

Generating BTs from task descriptions is a fundamental objective in the domain of automated planning and execution for robotics applications. 
The primary goal is to transform a given task description into an executable action plan structured as BTs.
This enables robots to perform complex tasks autonomously by following the logical flow dictated by the BTs.
\begin{equation}
\label{definition}
\small
    BT=G(task)
\end{equation}
As shown in Eq \ref{definition}, the BT generation task is defined as a function $G$ that maps a task description task onto a corresponding BT.
The BT generation task requires the following abilities: task understanding, task planning, specific behavior generation, as well as verification and validation of the generated BTs.

\textbf{Task Understanding} is significant for the BT generation since it is the first step in structuring the task into a BT. 
Without a clear understanding, it is impossible to create BTs to successfully complete the specified tasks and meet the requirements of users.
Extracting relevant information for generation, such as task objectives, robot hardware capabilities, and environmental constraints, is particularly challenging when the task description is provided in unconstrained natural language form. 

\textbf{Task Planning} decomposes the complex task into multiple subtasks and organizes them in BT generation.
Task planning requires converting abstract semantic instructions into detailed actionable sub-tasks and organizing them in the correct order.
Specifically, it decomposes high-level goals into simpler manageable sub-tasks and organizes their control and information flows into the structure of a BT. 

\textbf{Specific Behavior Generation} lays the groundwork for creating a BT by defining the actions and decision rules that the system will use to interact with its environment. 
It is a critical step that influences the functionality, reliability, and effectiveness of the resulting BT. 
Without properly generated behaviors, a BT would lack the necessary elements to operate or achieve its intended goals.

\textbf{V\&V} in the BT generation is critical for creating systems that are dependable, safe, and effective. 
It prevents the implementation of flawed behaviors and guarantees that the designed behaviors are within the system's capability to carry out, ultimately contributing to the successful application of BTs in complex, real-world tasks.
    
LLMs have shown promise in the BT generation.
The advent of LLMs represents a recent milestone in machine learning, showcasing immense capabilities in natural language processing tasks and textual generation \cite{zhao2023survey}.
\textit{For Task Understanding}, LLMs demonstrate an exceptional ability in understanding natural languages as well as in the creation of new content that is coherent, contextually relevant, and appropriate.
\textit{For Task Planning}, LLMs are equipped with intricate mechanisms for strategic planning, dynamically generating multi-step action sequences and adapting them to evolving scenarios.
\textit{For Behavior Generation}, although LLMs do not directly create BTs, they can play a supportive role in the behavior generation process and bridge the gap between human communication and machine-executable behavior sequences. 
\textit{For Verification and validation}, recent scientific reports indicate that LLMs have the potential to function as comprehensive world models, simulating complex environments and scenarios with a high degree of fidelity, which could provide robust frameworks for testing the reliability and performance of AI-driven systems before they are developed in real-world applications.

Applying LLMs to the BT generation tasks still faces multiple challenges.
Obtaining an effective BTGen model based on LLMs can be divided into three steps: \textit{training, developing} and \textit{V\&V}. 
We summarize these challenges from these three steps as follows. 

First, in terms of training the BTGen model, the ability of BT generation involves key considerations about the interaction between model performance and its constituent elements. 
We simply encapsulate this ability, denoted as $\mathcal{A}_{\text{BTGen}}$, in Eq \ref{model_data}.
\begin{equation}
    \mathcal{A}_{\text{BTGen}}=\mathcal{M}\oplus \mathcal{D}
    \label{model_data}
\end{equation}
According to Eq \ref{model_data}, three items are included: the foundation LLM $\mathcal{M}$, the datasets used to train $\mathcal{D}$, and the training pipeline $\oplus$.
Thus, three challenges remain to be addressed in training LLMs for the BT generation.
\textbf{Challenge 1:} How to select a proper foundation LLM for the BT generation?
Many foundation LLMs with billions of parameters have been proposed, such as GPT \cite{OpenAI2023GPT4TR}, LLaMA \cite{touvron2023llama_a}, and GLM \cite{du2021glm}.
The BTGen model can be trained on these foundation LLMs incrementally rather than being trained from scratch when computing resources are limited.
However, how to choose an appropriate foundation LLM based on the abilities required for the BT generation needs to be addressed.
To the best of our knowledge, there is limited research in this area.
\textbf{Challenge 2:} How to generate a high-quality BT dataset?
A growing body of work emphasizes the impact of data quality on the performance of LLMs \cite{li2023starcoder, di2023codefuse}.
Data quality significantly impacts the performance of LLMs in terms of their abilities to understand language and provide reliable outputs.
A high-quality dataset is integral to enhancing both the efficiency and the performance of the BT generation process.
However, how to prepare a high-quality BT dataset for training a BTGen model has received limited attention. 
\textbf{Challenge 3:} How to develop a training pipeline for the BT generation?
Moreover, optimizing the interaction between foundation LLMs and data quality is important for enhancing the ability of AI systems \cite{goodfellow2016deep, zhang2021understanding}. 
This balance between model complexity and data quality acts as a guiding principle to maximize the learning effects of the model on collected BT datasets.
A good training pipeline that effectively aligns the BT generation with LLM original abilities is needed.

Second, the BTGen model needs to be employed in real-world scenarios and address practical challenges.
The BTGen model still struggles with a series of problems, such as hallucination, data bias, lack of domain knowledge, and explanation after the training process.
The next challenge is \textbf{Challenge 4:} How to develop the BTGen model in real-world scenarios?
A good development pipeline tailored to BT generation integrates advanced strategies, such as prompts and agents, to address and attenuate these prevalent challenges inherent in LLMs, paving the way to generate BTs with a higher degree of sophistication and operational reliability.

\begin{figure*}[htbp]
    \centering
    \includegraphics[width=5in]{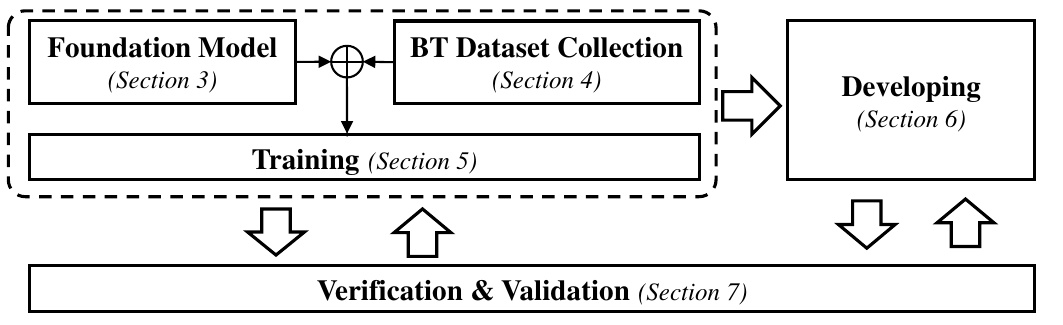}
    \caption{The overview structure of our paper.}
    \label{overview}
\end{figure*}

Lastly, measuring the effectiveness of the BTGen model after training and developing is crucial.
Thus, the last challenge is \textbf{Challenge 5:} How to verify and validate the BTGen model?
Many researches have reported that LLMs cannot generate correct outcomes when faced with complex tasks.
Thus, an effective V\&V pipeline is imperative for the BT generation.
By instituting rigorous evaluation protocols and testing methodologies, the V\&V pipeline can guarantee that the BTs not only operate effectively within their intended operational parameters but also conform to stringent quality standards.
Consequently, the incorporation of a comprehensive V\&V pipeline thus serves as a cornerstone for the utilization of BTGen model in producing dependable BTs.

In this paper, we explore the use of LLMs to enhance the BT generation and offer insights and perspectives on overcoming the aforementioned challenges.
The main structure of our paper is shown in Figure \ref{overview}, including the training phase, developing phase, and verification and validation phase.
Specifically, the training phase is further divided into foundation LLM selection, BT dataset collection, and the training pipeline.
We introduce each component that addresses the above-summarized challenges in each section in detail.

\section{Foundation Models}\label{sec:foundation_models}
\label{sec:basemodel}

\begin{figure*}[htbp]
    \label{gpt_series}
    \centering
    \includegraphics[width=6in]{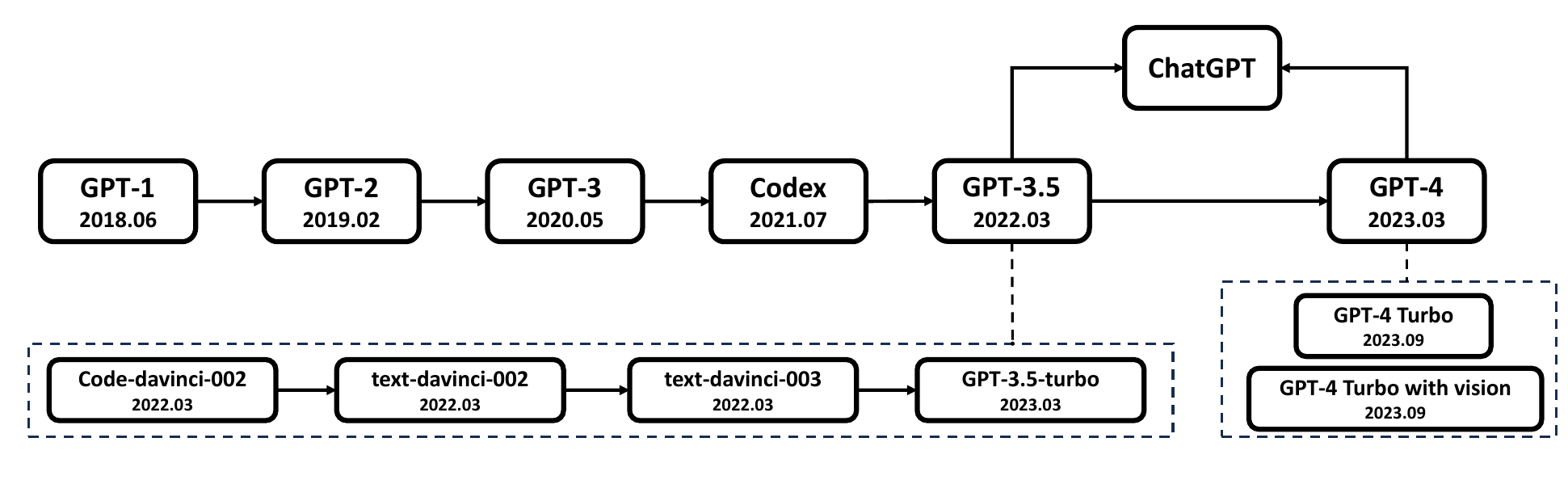}
    \caption{A brief illustration for the evolution of GPT-series models \cite{zhao2023survey}. 
    }
    % \vspace{-0.5cm}
\end{figure*}

A foundation model is a model that is trained on broad data (generally using self-supervision at scale) and can be adapted (e.g., fine-tuned) to a wide range of downstream tasks \cite{bommasani2021opportunities}. These models are based on deep neural networks and self-supervised learning, both of which have existed for decades. However, the recent scale and scope of foundation models have pushed the boundaries of what is possible. Foundation models are typically huge, with billions or even trillions of parameters, which allows them to learn complex language patterns and perform tasks that would be difficult or impossible for smaller models. In recent years, several foundation LLMs have been proposed by different organizations, each with its own characteristics. We briefly summarize and introduce three representative foundation LLMs: GPT-series, LLaMA-series, and GLM-series.

\textbf{GPT-Series: }
The GPT (Generative Pre-trained Transformer) series of LLMs developed by OpenAI\footnote{https://openai.com/} has undergone significant development over the years. The family of GPT-series models is briefly shown in Figure \ref{gpt_series}. The first two initial models are GPT-1 \cite{scao2022language} and GPT-2 \cite{radford2019language}, which can be considered as the foundation for more powerful models subsequently, such as GPT-3 \cite{dong2022survey} and GPT-4 \cite{OpenAI2023GPT4TR}. GPT-1 was introduced in June 2018 and developed based on a generative, decoder-only Transformer architecture with 117 million parameters. It adopts a hybrid approach of unsupervised pretraining and supervised fine-tuning.  GPT-1 \cite{scao2022language} set up the core architecture for the GPT-series models and established the underlying principle of modeling natural language text, i.e., predicting the next word. GPT-2 was introduced in February 2019 and increased the parameter scale to 1.5B, trained with a large webpage dataset. As claimed in the paper of GPT-2 \cite{radford2019language}, it seeks to perform tasks via unsupervised language modeling, without explicit fine-tuning using labeled data. Based on GPT-2, GPT-3 \cite{dong2022survey} released in 2020 demonstrated a significant capacity leap by scaling the generative pre-training architecture to an even larger size of 175B parameters. GPT-3 can be viewed as a remarkable landmark in the journey from PLMs to LLMs. It empirically proved that scaling the neural networks to a significant size can lead to a huge increase in model capacity. The conversation model ChatGPT\footnote{https://openai.com/chatgpt} was released by OpenAI in November 2022, based on the GPT models (GPT-3.5 and GPT-4).  ChatGPT exhibits superior capacities in communicating with humans, possessing a vast store of knowledge, skill at reasoning on mathematical problems, accurately tracing the context in multi-turn dialogues, and aligning well with human values for safe use. GPT-4 \cite{OpenAI2023GPT4TR}, released in March 2023, extends the text input to multimodal signals with more parameters. Overall, GPT-4 has stronger capacities in solving complex tasks than all former GPT-series models, showing a large performance improvement on many evaluation tasks.  
    
\textbf{LLaMA-Series:} LLaMA (LLM Meta AI) \cite{touvron2023llama_a} is a family of LLMs released by Meta AI starting in February 2023. LLaMA foundational models are trained on a dataset with 1.4 trillion tokens, drawn from publicly available data sources. Four model sizes were trained: 7, 13, 33, and 65 billion parameters. LLaMA uses the transformer architecture with minor architectural differences.  Compared to GPT-3, LLaMA uses the SwiGLU \cite{shazeer2020glu} activation function instead of ReLU, rotary positional embeddings \cite{su2023roformer} instead of absolute positional embedding, and root-mean-squared layer-normalization instead of standard layer-normalization. LLaMA 2 \cite{touvron2023llama_b} is trained on a dataset with 2 trillion tokens and increases the context length from 2K tokens to 4K tokens. Three model sizes, 7, 13, and 70 billion parameters, are released. LLaMA 2-Chat is additionally fine-tuned on 27,540 prompt-response pairs created for this project, which performs better than larger but lower-quality third-party datasets. For AI alignment, reinforcement learning with human feedback (RLHF) is used with a combination of 1,418,091 Meta examples and seven smaller datasets.
    
\textbf{GLM-Series:} GLM (General Language Model) \cite{du2021glm} is pretrained with an autoregressive blank-filling objective. It randomly blanks out continuous spans of tokens from the input text, following the idea of auto-encoding, and is trained to sequentially reconstruct the spans, following the idea of autoregressive pertaining. ChatGLM-6B is a conversational language model that supports bilingual question-answering in Chinese and English. ChatGLM-6B adopts the same model architecture as GLM-130B. As of July 2022, GLM-130B \cite{zeng2022glm} has only been trained on 400 billion tokens with a 1:1 ratio of Chinese to English. In contrast, ChatGLM-6B utilizes a larger training dataset with up to 1 trillion tokens, consisting solely of Chinese and English text in a 1:1 ratio.

\begin{table*}[htbp]
  \centering
  \renewcommand\tabcolsep{2pt}
  \footnotesize
  \label{tab:basic_abilities}%
  \caption{Abilities of foundation LLMs related to the BT generation.}
    \begin{tabularx}{\textwidth}{XXXX}
    \hline\hline
    \textbf{Categories} & \textbf{Abilities} & \textbf{Description} & \textbf{BT Generation} \\
    \hline
    \multirow{2}[0]{*}{\textbf{Natural-Language-Related}} 
    & Natural language understanding
    & Comprehend and interpret natural language.
    & Understand the meaning of actions, entities, and intentions of the BT generation tasks. \\
    \cmidrule{2-4}
    & Natural language generation
    & Generate human-readable texts.
    & Achieve better human-computer interaction and return natural language texts to explain the process in the BT generation for humans. \\
    \cmidrule{2-4}
    & In-context learning
    & Learn and perform new tasks by observing a few examples.
    & Learn novel knowledge from few-shots demonstrations, especially in unseen scenarios of the BT generation tasks. \\
    \cmidrule{2-4}
    & Instruction following
    & Understand and carry out instructions or commands accurately and effectively.
    & Align with the human intentions in the BT generation. \\
    \hline
    \multirow{2}[0]{*}{\textbf{Reasoning-Related}} 
    & Commonsense reasoning
    & Understand and use general knowledge that is common to the most people in their daily lives.
    & Make reasonable assumptions and inferences in the absence of specific information.\\
    \cmidrule{2-4}
    & Logical reasoning
    & Using logical principles and rules to arrive at valid conclusions or make sound judgment.
    & Understand and make the transition of actions and states in BTs. \\
    \cmidrule{2-4}
    & Planning
    & Decompose a potentially complex task into subtasks.
    & Decompose the complex BT generation tasks into several simple subtasks that are easy for LLMs to implement. \\
    \hline
    \textbf{Tool-Related} 
    & Retrieval
    & Enable AI to find the proper knowledge from database based on the given query conditions.
    & Find corresponding tools or knowledge for BT generation tasks.\\
    \cmidrule{2-4}
    & Tool manipulation
    & Enable AI to manipulate the existing tools to get the desired outputs.
    & Use specialized tools, such as the node libraries and visual detectors for BT generation tasks.\\
    \hline
    \textbf{Code-Related} 
    & Code generation
    & Generating program codes.
    & Generate the BT codes that are executable and meet the requirements of users. \\
    \cmidrule{2-4}
    & Code explanation
    & Return the concrete meanings of the code program.
    & Explain the BT codes in natural languages for humans. \\
    \cmidrule{2-4}
    & Code translation
    & Translate the codes from one language to other language.
    & Translate BT codes between different code languages for different goals.\\
    \hline\hline
    \end{tabularx}%
\end{table*}%

These foundation models exhibit strong capacities in the area of natural language after being trained on large unlabeled corpus datasets.
However, the BT generation tasks are more complex compared to text generation.
As foundation LLMs, certain basic abilities are needed for BT generation tasks.
We categorize the basic abilities of foundation LLMs for BT generation into four principal domains: natural language-related abilities, reasoning-related abilities, tool-related abilities, and code-related abilities, as shown in Table \ref{tab:basic_abilities}.

\subsection{Natural-Language-Related Abilities}

Natural language-related abilities are the core abilities of human linguistic competence.
These abilities include natural language understanding, natural language generation, in-context learning, and instruction following.
\textit{Natural language understanding} refers to the ability to understand natural language text and extract useful information for downstream tasks. 
Common tasks in natural language understanding include sentiment analysis, text classification, natural language inference, and semantic understanding.
\textit{Natural language generation} evaluates the capabilities of LLMs in generating specific texts, including tasks such as summarization, dialogue generation, machine translation, question answering, and other open-ended generation tasks.
\textit{In-context learning (ICL)} is a paradigm that allows language models to learn tasks given only a few examples in the form of demonstrations \cite{dong2022survey}.
With ICL, the pre-training and utilization of LLMs converge to the same language modeling paradigm: pre-training predicts the following text sequence conditioned on the context, while ICL predicts the correct task solution, which can also be formatted as a text sequence, given the task description and demonstrations.
\textit{Instruction following} enables LLMs to understand and carry out instructions or commands accurately and effectively.
With instruction following, LLMs can align with the intentions of humans, follow task instructions for new tasks without using explicit examples, and improve their generalization ability.

\textbf{Related to BT Generation}:
These natural language-related abilities play a crucial role in the BT generation as they enable LLMs to understand the task description, environmental information, and the desired objective states. 
Under prompts involving a few examples in similar scenarios and specific instructions, the models can effectively comprehend and interpret the provided instructions, ensuring a clear understanding of the task requirements. 
Moreover, these abilities empower the models to generate human-readable and explainable actions, allowing for transparent and interpretable decision-making processes. 
This not only enhances the model's overall performance but also promotes trust and understanding between the model and its users.

\subsection{Reasoning-Related Abilities}

Complex reasoning encompasses the capacity to comprehend and effectively employ supporting evidence and logical frameworks to deduce conclusions or facilitate decision-making. 
By analyzing the reasoning abilities required for BT generation, we believe that the three most crucial reasoning abilities currently needed for BT generation are commonsense reasoning, logical reasoning, and planning.
\textit{Commonsense reasoning} is a fundamental ingredient of human cognition, encompassing the capacity to comprehend the world and make decisions. 
This cognitive ability plays a pivotal role in developing natural language processing systems capable of making situational presumptions and generating human-like language.
\textit{Logical reasoning} is the ability to examine, analyze, and critically evaluate arguments as they occur, holding significant importance in understanding.
The chain of thoughts (CoT) \cite{wei2022cot} is a well-known logical reasoning technique that solves problems or accomplishes tasks through a series of intermediate steps and logical operations.
With the CoT prompting strategy, LLMs can solve many tasks by utilizing the prompting mechanism that involves intermediate reasoning steps for deriving the final answer.
\textit{Planning} decomposes a potentially complex task into simpler subtasks that the LLM can solve more easily by itself or using tools according to \cite{mialon2023augmented}.
Planning involves developing a course of actions (policy) to execute tasks, which takes the agent to a desired state of the world. 

\textbf{Related to BT Generation}:
These three capabilities synergistically combine and collaborate in the intricate process of the BT generation based on LLMs. 
Commonsense reasoning serves as the foundation, allowing the model to grasp the task context and apply contextual knowledge to its decision-making. 
Logical reasoning ensures the coherence and rationality of the reasoning process, enabling the model to make sound judgments and logical connections between different components of BTs. 
Planning facilitates the effective execution of the task by devising efficient strategies to accomplish the desired objectives.
Therefore, LLMs equipped with these capabilities possess a profound understanding of the task at hand, ensuring the generation of complex BTs.

\subsection{Tool-Related Abilities}

Tool-related abilities refer to the capabilities of foundation models to manipulate tools, leading to more potent and streamlined solutions for real-world tasks.
These abilities enable LLMs to interact with the real world, such as manipulating search engines \cite{nakano2021webgpt}, shopping on e-commerce websites \cite{yao2022webshop}, planning in robotic tasks \cite{huang2022language, huang2022inner}.
We summarize the tool-related abilities into retrieval and tool manipulation.
\textit{Retrieval} refers to the model's ability to efficiently and accurately retrieve query-related information from a knowledge library. 
This includes retrieving relevant tools, domain knowledge, and other pertinent information based on the given query conditions. 
The model should be capable of effectively searching and retrieving the most relevant and useful information to assist in addressing the query or task at hand.
\textit{Tool manipulation} refers to the ability of LLMs to utilize various tools or software to perform specific tasks. 
It enhances and expands the performance of models, leading to more potent and streamlined solutions for real-world tasks, bridging the gap between its language generation capabilities and practical scenarios.

\textbf{Related to BT Generation}:
The tool-related abilities are crucial for complex BT generation tasks as they allow leveraging the functionalities and expertise of existing tools and knowledge. 
Given the complexity and domain-specific nature of the BT generation tasks, utilizing domain-specific tools simplifies the process and significantly enhances the quality of BTs.
These tools offer specialized functionalities like visual editors, code analysis, and debugging, greatly assisting in creating and validating the BT generation. 
By integrating with these tools, LLMs can generate effective BTs while leveraging domain-specific knowledge.

\subsection{Code-Related Abilities}
Code-related abilities play a key role in ensuring the correctness and executability of BTs.
We summarize three main abilities: code generation ability, code explanation ability, and code translation ability.
\textit{Code generation ability} is the core ability in the code-related category and aims to make LLMs automatically generate a program that satisfies a given natural language requirement.
It requires LLMs to generate code snippets, complete functions, or even entire programs based on given prompts or specifications.
To enable LLMs to acquire code generation ability, mainstream LLMs are trained on large code repositories from GitHub, competitive programming websites, and public benchmarks.
They can be trained from scratch or based on existing LLMs.
CodeGen \cite{nijkamp2022codegen}, StarCoder \cite{li2023starcoder}, and CodeGeeX \cite{di2023codefuse} are representative works that are trained from scratch based on a decoder-only transformer architecture.
They are pre-trained in an autoregressive manner on a large corpus of code, resulting in significant training costs.
In contrast, some works perform incremental training based on existing LLMs to obtain code generation capabilities.
Codex \cite{chen2021evaluating} is incrementally trained based on GPT-3, and Code LLaMA \cite{roziere2023codellama} is incrementally trained on LLaMA-2.
\textit{Code explanation ability} refers to describing and clarifying what a piece of code does, how it works, its purpose, and its logic in a human-understandable way. 
The goal of code explanation is to make the code maintainable and understandable for anyone who might read it in the future, including the original author. 
\textit{Code translation ability} refers to converting code written in one programming language into another programming language while preserving the original program's functionality and logic under different requirements.
Translating code often requires significant refinement and testing to ensure it behaves as intended and follows the conventions of the target language.

\textbf{Related to BT Generation}:
Since BTs are represented in the form of code, the code-related abilities play a pivotal role in the BT generation tasks, ensuring the generation of executable and human-desired codes.
These abilities facilitate the translation of abstract instructions into actionable code that can be readily executed by agents or systems, explain the codes for developers in natural language, and enhance the transparency and interpretability of the generated BTs.

\section{BT Dataset Collection}\label{sec:bt_data_collect}
As highlighted in Section \ref{sec:method}, the role of data is pivotal in the development of proficient LLMs. In the context of generating BTs, a high-quality dataset is integral to enhancing both the efficiency and performance of the BT generation process.

\subsection{Dataset Composition}
We commence by outlining the structure and format of the dataset required for training our LLM-based BT generation. Given that our approach leverages the capabilities of LLMs, it is essential to incorporate both structured and natural language elements within each data entry. We propose a novel dataset schema tailored for BT generation, encompassing five key components: \textit{Name}, \textit{Description}, \textit{XML Representation}, \textit{Nodes Metadata}, and \textit{Implementations}.

\begin{itemize}
    \item \textbf{Name}: This field specifies the identifier of the BT, which is meaningful in both a domain-specific and a natural language context. For instance, the name could reflect the type of robotic platform or the intended objective of the BT.
    \item \textbf{Description}: A comprehensive description of the BT's purpose is provided here, detailing aspects such as the robotic platform it is designed for, the task it aims to achieve, and any relevant environmental considerations.
    \item \textbf{XML Representation}: An XML-formatted representation encapsulates the structural and node-specific details of the BT, facilitating both human readability and machine parsability.
    \item \textbf{Nodes Metadata}: This part elucidates the functionality and operational logic of individual nodes within the tree, encompassing their respective names and roles.
    \item \textbf{Implementations}: This section delineates the concrete code implementations for the behavior encapsulated by each node, promoting an understanding of the practical application of the tree's logic.
\end{itemize}

To elucidate the proposed data format, we present an illustrative example in Table \ref{tab:xml_format}. 

\begin{table*}[htbp]
  \centering
  \caption{A piece of formatted BT data}
  \fontsize{10}{12}\selectfont
  \begin{tabularx}{\linewidth}{p{0.08\textwidth}X}
    \hline
    \textbf{Name} & \textbf{UAV\_Patrol\_Campsite} \\
    \hline
    \textbf{Desc.} & {\small The tree describes the UAV patrolling based on a predetermined route. If a suspicious target is detected, the UAV will warn the target, otherwise, the UAV will move to the next location in the route.} \\
    \hline
    \multirow{7}[2]{*}{\textbf{XML}} &
    \small \textless Fallback instance\_name = 'fallback\_node'/\textgreater \\
    & \small \quad \textless Sequence instance\_name = 'sequence\_node'/\textgreater \\
    & \small \quad \quad \textless Condition instance\_name = 'check-target\_detected'/\textgreater \\
    & \small \quad \quad \textless Action instance\_name = 'warn-target' /\textgreater \\
    & \small \quad \textless /Sequence\textgreater \\
    & \small \quad \textless Action instance\_name = 'move-to\_next-pos'/\textgreater \\
    & \small \textless /Fallback\textgreater \\
    \hline
    \multirow{3}[2]{*}{\textbf{Nodes}} & \small \textbf{check-target\_detected}: Check if any suspicious targets have been detected. \\
    \cline{2-2}  
    & \small \textbf{warn-target}: Warn the target. \\
    \cline{2-2}  
    & \small \textbf{move-to\_next-pos}: Move to the next location in the route. \\
    \hline
    \multirow{23}[6]{*}{\textbf{Impls}} & \small \textbf{check-target\_detected:} \\
    & \small \quad class \textbf{class} CheckTarget \\
    & \small \{ \\
    & \small \quad CheckTarget() \{ \}; \\
    & \small \quad \quad node\_status run() \\
    & \small \quad \quad \{ \\
    & \small \quad \quad \quad targetDetected = getStatus(target); \\
    & \small \quad \quad \quad \textbf{if} (targetDetected) \\
    & \small \quad \quad \quad \quad \quad \textbf{return} Success; \\
    & \small \quad \quad \quad \textbf{else} \\
    & \small \quad \quad \quad \quad \quad \textbf{return} Failure; \\
    & \small \quad \} \\
    & \small \} \\
    \cline{2-2}  
    & \small \textbf{warn-target:} \\
    & \small \quad class \textbf{class} Warn \\
    & \small \{ \\
    & \small \quad Warn() \{ \}; \\
    & \small \quad \quad node\_status run() \\
    & \small \quad \quad \{ \\
    & \small \quad \quad \quad \textbf{try} \\
    & \small \quad \quad \quad \quad \quad target = getTarget(); \\
    & \small \quad \quad \quad \quad \quad warning\_status = warn\_target(target); \\
    & \small \quad \quad \quad \quad \quad setStatus(warning\_status); \\
    & \small \quad \quad \quad \quad \quad \textbf{return} Success; \\
    & \small \quad \quad \quad \textbf{catch}(...) \\
    & \small \quad \quad \quad \quad \quad \textbf{return} Failure; \\
    & \small \quad \} \\
    & \small \} \\
    \cline{2-2}  
    % & \textbf{move-to\_next-pos:} $\cdots$ \\
    & \small \textbf{move-to\_next-pos:} ··· \\
    \hline
  \end{tabularx}
  \label{tab:xml_format}%
\end{table*}%

\subsection{Synthetic Data Generation}

The utilization of LLMs typically necessitates the availability of vast amounts of data. In many cases, such datasets are derived from human-generated sources, which present several challenges: they can be costly to gather \cite{Inbar_Oren_}, are often difficult to obtain \cite{Naman_Bansal_}, may suffer from limitations in scope and diversity \cite{Alon_Shoshan_}, or, in some instances, are simply not available \cite{Vietri2020,Terrance_Liu_}. Within the domain of BT generation, these issues are exacerbated due to the dearth of BT-specific datasets, which tend to be scattered across various platforms such as paper-based prototypes, gaming environments, and robotic applications. The scarcity of BT data hinders the effective training or fine-tuning of LLMs.

Yet, recent advancements have shown that synthetic data, generated by models themselves, can yield performance on par with, if not superior to, that of human-generated counterparts \cite{singh_beyond_2023}. Synthetic data provides a viable solution to circumvent the limitations posed by human-generated datasets, enhancing sample efficiency \cite{Inbar_Oren_}, mitigating the shortage of training data, and reducing the cost associated with data collection. Therefore, the automatic construction of high-quality synthetic datasets emerges as an effective strategy to quickly expand the volume of available training data. Our focus will be on delineating the methodologies for generating synthetic data pertinent to BTs.

Approaches to data synthesis can generally be classified into two categories: those based on domain information and those leveraging LLMs. Domain-based methods may utilize features inherent to the data, exploit domain-specific knowledge \cite{Ruohao_Guo_}, or employ domain-centric tools \cite{Matt_Maufe_}. On the other hand, LLM-based synthesis capitalizes on the ability of LLMs to generate contextually relevant content through specific instructions \cite{Libing_Zeng_}.

\subsubsection{Learning-Based Methods}

Learning-based methods for BT generation demonstrate superior adaptability by effectively representing strategic behaviors within BT structures \cite{Chen2023}. These methods leverage large amounts of data to learn behavior patterns and strategies, enabling agents to adapt more effectively across diverse tasks and environments. By automatically discovering and capturing strategic behaviors, learning methods empower agents to make adaptive decisions in complex environments. Furthermore, through iterative training, these methods can refine the generated BTs based on its actual effectiveness, ensuring better alignment with desired goals and requirements. Notable examples of these methods include RL-based approaches \cite{Chen2023, Ahmad2023}, evolution-based techniques \cite{Partlan2022, Iovino2023}, and demonstration-based methods \cite{french2019learning, WOS:000972696000033}.

\textbf{RL-Based BT Generation}. 
Reinforcement learning (RL) has proven effective in the generation of BTs. The process entails defining state and action spaces, creating reward functions \cite{Ahmad2023}, establishing RL models \cite{Zhu2019}, and determining appropriate representations for BTs \cite{Chen2023}. Notably, RL has demonstrated its capability to learn both the structure and parameters of BTs, making them applicable in dynamic settings \cite{WOS:000766992600188}. Moreover, hierarchical RL techniques have been employed to decompose tasks into subtasks, each with its localized strategy \cite{WOS:001050787900006}.
    
\textbf{Evolution-based BT Generation}. 
Evolution-based methods, centered around the genetic algorithm, play a crucial role in BT generation. These methods emulate biological evolution by employing selection, crossover, and mutation processes to optimize BTs based on fitness evaluation outcomes \cite{Nicolau2017}. In practical implementation, populations are initialized, renewed, and termination criteria are established \cite{Partlan2022}. Genetic programming is one approach used for BT generation, which involves defining encoding methods \cite{Iovino2023}. Another approach is grammar evolution, where grammatical rules are specified to constrain the search space of evolutionary algorithms, simplifying problem-solving \cite{Neupane2022}.
    . Genetic programming has proven effective in handling highly uncertain environments during BT generation \cite{iovino2021learning}, while grammar evolution aids in constraining the search space and streamlining the evolutionary process \cite{WOS:000896394100024}.
    
\textbf{Demonstration-Based BT Generation}. 
Learning from demonstrations allows the acquisition of complex, multi-step tasks through observation of expert-provided examples \cite{french2019learning}. Such a method captures the decision-making pathways and rule sets within these demonstrations, extracting the underlying structures and regulations to form BTs. This approach has enabled the generation of BTs by observing human behaviors \cite{WOS:000537388400001} and has led to the development of semi-automatic optimization methods for adjusting expert-constructed BTs \cite{WOS:000972696000033}.

\subsubsection{Planning-Based Methods}

Planning-based methods offer an intuitive and interpretable framework for BT generation. These methods are grounded in the principles of automated planning and have been explored extensively in the literature \cite{colledanchise2017synthesis,lan2019autonomous,rovida2017extended,colledanchise2019towards}. Colledanchise et al. introduced a novel reactive planning algorithm that facilitates the automatic generation and dynamic updating of BTs, effectively integrating the modular and reactive nature of BTs with the systematic synthesis process of automated planning \cite{colledanchise2019towards}. We categorize planning-based methods into three distinct types: logic-based, hierarchical, and active planning methods.

\textbf{Logic-Based Method}
Approaches such as linear temporal logic (LTL) have been utilized to synthesize BTs \cite{colledanchise2017synthesis,lan2019autonomous}, providing a formal mechanism to define complex behaviors. For instance, \cite{WOS:000981889200003} employs decision tree learning combined with logical decomposition techniques to generate BTs from execution traces of previously manually designed plans. Additionally, a top-down divide-and-conquer strategy is proposed in \cite{WOS:000799032800006}, which simplifies complex F-LTL formulas into manageable sub-formulas that facilitate the automatic BT generation.

\textbf{Hierarchical Method}
Hierarchical methods utilize task networks to structure BTs. A state space representation can be incorporated to ensure the soundness and completeness of BTs, enabling robots to handle all solvable external disturbances \cite{WOS:000680423506019}. \cite{WOS:000687918700007} integrates hierarchical planning to construct BTs systematically. Coordination among intelligent vehicles is achieved using a hierarchical auction algorithm in \cite{WOS:000933901200001}, while state-aware hierarchical BTs are developed to support post-action decision-making, preferences, and local priorities within robotic applications \cite{WOS:000992458200045}. Francesco et al. advocated for the use of hierarchical task networks optimized for execution time efficacy in BT creation \cite{rovida2017extended}.

\textbf{Active Planning Methods}
Active planning methods merge environmental perception with planning capabilities to address the dynamics of action execution contexts. A method incorporating both environmental awareness and planning to infer context and automatically generate BTs is presented in \cite{WOS:000885903300165}. Hybrid active planning, which allows BTs to dynamically adjust in response to environmental changes, is discussed in \cite{WOS:001031697300012}. Further, active reasoning is leveraged in \cite{WOS:000910157700001} through the introduction of new leaf node types that specify desired states and determine actions online to achieve those states. Finally, strategies for generating BTs in partially observable environments, aimed at mitigating uncertainty within conditional nodes, are explored in \cite{WOS:000724145801068}.

Nevertheless, it is imperative to acknowledge the inherent limitations of planning-based methods. Their dependency on pre-defined knowledge constrains their ability to adapt to unforeseen circumstances or environments where explicit rules may not exist. This reliance on existing domain knowledge poses challenges when encountering novel scenarios, necessitating further research to enhance the robustness and generality of BT generation in dynamic and unpredictable settings.

\subsubsection{LLM-Based Methods}

The advent of large language models has catalyzed innovative approaches in BT generation by harnessing the generative and semantic comprehension capabilities of such models. The advanced linguistic processing prowess of LLMs facilitates the crafting of detailed descriptions and definitions, thereby enabling the generation of high-quality BTs from task narratives or constraints. This method encompasses various stages including task description completion, task decomposition, node selection, BT structuring, preliminary BT validation, and iterative BT refinement.

Among the notable applications, \cite{WOS:000909405303096} encodes a BT into a numerical vector while preserving both the semantic content of task descriptions and the structural intricacies of hierarchical task arrangements through pre-trained language embeddings and node aggregation mechanisms. In another study, \cite{Cao2022} introduces an embedding methodology for BTs that accepts as input a BT associated with individual tasks and outputs the corresponding numerical representation.

\subsubsection{Alternative Methods}

In the paper by Gonzalo Florez-Puga \cite{Gonzalo_Florez-Puga_2009}, Case-based Reasoning (CBR) is employed to facilitate the retrieval and adaptation of existing BTs, allowing for the reuse of previously developed BTs. This approach enhances efficiency and promotes knowledge transfer. El-Ariss \cite{WOS:000702263300122} proposes a novel extension to traditional BTs that enables them to recall and utilize information from previously executed subtrees, leveraging historical context to make informed decisions. This enhancement improves the sophistication and reusability of BTs, enabling them to adapt and improve based on past experiences. Umeyama \cite{WOS:000799480000104} constructs an action node graph using geospatial data, which is then automatically converted into a nuanced BT, enhancing its capabilities and effectiveness for context-aware decision making. Scheide \cite{WOS:000765738803108} formalizes the search space for BTs as a grammar, allowing the BT structure to be derived through exploration using the Monte Carlo tree search algorithm. This formalization provides a systematic and efficient approach to generating BTs, improving the overall effectiveness of the generation process. Lastly, Gao \cite{WOS:000812286900309} introduces a three-phase BT generation algorithm that includes construction, simulation, and online re-planning, enabling the generation of adaptive BTs that can dynamically adjust their behavior based on the current situation, enhancing their flexibility and robustness.

\subsection{Node Library}

The Node Library is pivotal for assuring that the execution of BTs fulfills its purpose effectively and translates abstract task directives into operational code. This library is a cornerstone in refining task descriptions and furnishing leaf nodes that are both adaptable to various software and hardware environments and designed with scalability in mind. 

To bolster data synthesis, it is essential to curate a Node Library that offers a repository of reusable node types crucial for the assembly of BT data. The library's design emphasizes ease of use and modularity, enabling straightforward adoption and extension of nodes to suit a variety of developmental needs. Moreover, by modularizing the development workflow, the Node Library streamlines the design process, leading to improved productivity.

Within the context of BTs, nodes are generally partitioned into control nodes, which govern the BT's execution flow, and leaf nodes, which enact precise condition assessments and action implementation. Specifically: \textbf{Control Nodes} dictate the logical structure and sequencing within the BT. \textbf{Leaf Nodes} include \textit{Condition Nodes} and \textit{Action Nodes}. \textit{Condition Nodes} that appraise logical states and navigate the BT along different branches based on those evaluations. \textit{Action Nodes} that initiate distinct operations capable of altering environmental states.

The symbiosis of condition nodes and action nodes under the auspices of control nodes allows BTs to adapt to and make informed decisions in response to environmental contingencies, thus achieving sophisticated behavioral patterns. Our Node Library's primary aim is to facilitate robust implementations of these leaf nodes.

For each node, the following attributes are meticulously curated:

\begin{itemize}
    \item \textbf{Type}: A dichotomy encompassing condition nodes and action nodes typifies node classification.
    \item \textbf{Name}: Descriptive naming conventions enhance the readability and accessibility of the node library, fostering user-friendly interaction.
    \item \textbf{Description(Desc.)}: An elaborate exposition of the node's function delineates its objectives, behaviors, parameters (both input and output), and its contextual role within the BT framework. Clarity in function description is instrumental for developers and LLMs alike, ensuring appropriate node deployment in BT generation.
    \item \textbf{Implementation (Impl.)}: This constitutes the node's operational codebase, which includes simulation verification scripts and real-world operational code. Rigorous testing in simulated environments precedes deployment, validating the node's efficacy in fulfilling intended tasks and confirming its practical problem-solving capabilities in authentic settings.
\end{itemize}

\subsection{Our Method for Data Generation}

\begin{figure*}[htb]
    \centering
    \includegraphics[width=0.8\textwidth]{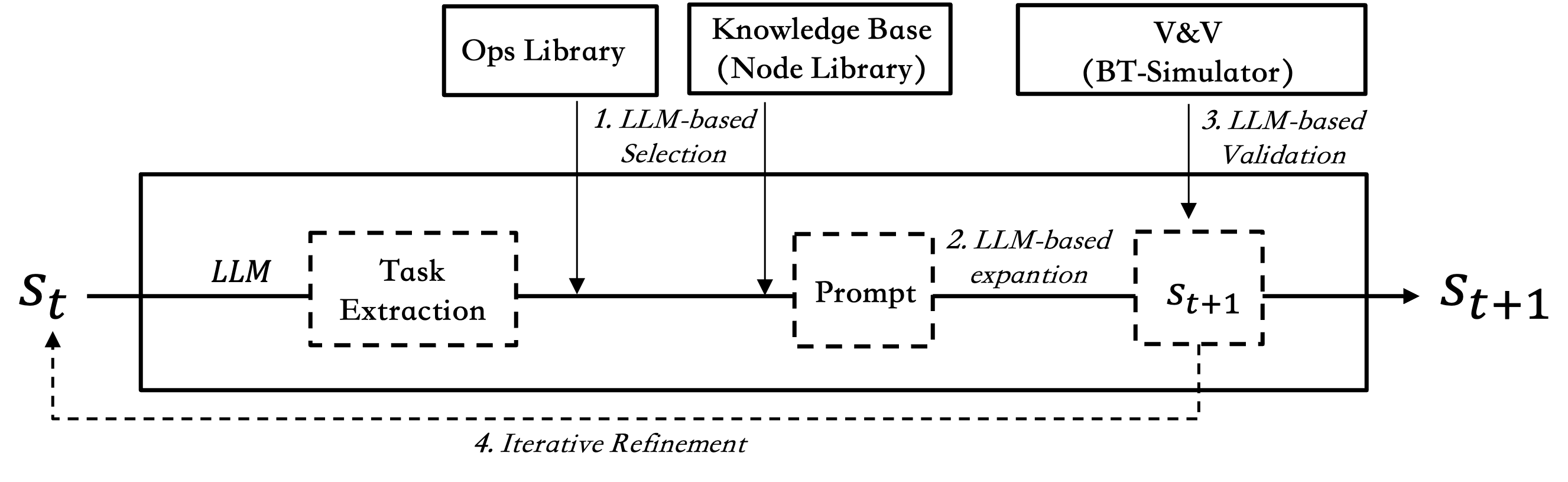}
    \caption{Illustration of the BT Generation method utilizing a framework inspired by Monte Carlo Tree Search (MCTS) for generating BTs. Each state represents a BT, with transitions informed by selection, expansion, validation, and iterative refinement processes. The approach integrates a language model for decision-making and node expansion.}
    \label{fig:btgen_mcts}
\end{figure*}

Despite LLMs demonstrating significant promise in BT generation, their application to domain adaptation is still an emerging field. The integration of BTs with natural language directives to guide agents in learning and executing complex tasks is explored in \cite{Suddrey2022}, with potential extensions to novel contexts.

\textbf{Hierarchical Generation with LLM Method}
Investigating automatic task generation based on BTs, \cite{Cao2023} delineates a stage-step prompting technique tailored for hierarchically structured robot tasks. Sequentially Ordered Tasks (SOT)\cite{ning2023sot} pioneer a strategy whereby LLMs draft a skeletal framework for responses before conducting parallel API queries or collective decoding to flesh out each element of the skeleton simultaneously.
    
\textbf{Reasoning with LLM Method}
Chain-of-Thought (COT)\cite{wei2022cot} empowers LLMs by articulating reasoning processes step-by-step, thereby augmenting the models' inferential capabilities. Expansion of Thoughts (XOT)\cite{ding2023thoughts} employs reinforcement learning coupled with Monte Carlo tree search methodologies to navigate thought processes, enriching LLMs with external domain expertise during problem-solving phases.
    
\textbf{Constrained Selection with LLM Method}
\cite{lykov2023llmbrain} presents an autonomous robotic control approach predicated on an LLM fine-tuned from the Stanford Alpaca 7B architecture, specializing in generating robot BTs from textual descriptions using a predefined node library.
    
\textbf{Retrieval with LLM Method}
In realms demanding specialized knowledge, database integration can elevate the expertise of LLMs. Introducing programmatic mapping layers bridges the gap between general-purpose and domain-specific vernaculars.

The emergence of LLMs has sparked new and innovative approaches in generating BTs by leveraging the generative and semantic comprehension capabilities of these models. The advanced linguistic processing capabilities of LLMs enable the creation of detailed descriptions and definitions, facilitating the generation of high-quality BTs from task narratives or constraints. This method involves several stages that collectively contribute to the generation process, such as  task description completion, task decomposition, node selected, structure design, verification and verification, reflection, etc.

The proposed methodology involves synthesizing synthetic data through interaction with Large Language Models (LLMs), guided by carefully structured prompts. We prioritize the fine-tuning phase in this work but also acknowledge the potential of investigating pre-training strategies in future research. The crux of effective fine-tuning lies in the construction of question and answer pairs; however, existing datasets often present challenges such as being incomplete or having an imbalance between questions and answers. LLMs excel in filling these gaps by producing the missing elements—whether that involves generating questions from given answers or vice versa. Contemporary tactics for LLM-based synthetic data generation are predominantly categorized into two streams: example imitation and data-driven information extraction.

To generate synthetic BTs, we leverage methods such as self-instruct \cite{wang2023selfinstruct} and evol-instruct \cite{xu2023wizardlm}, which draw upon an extensive problem seed pool. These methodologies integrate the seed pool data with foundational prompts to create new problem instances. The quality and diversity of the seed pool substantially affect the caliber of the resulting queries. Our focus is on formulating robust inputs for task-solving data, considering that solutions to certain problems are already accessible within established knowledge bases. Utilizing an LLM enables us to infer the corresponding problems grounded in these known solutions. A comprehensive seed data repository is constructed by analyzing the vast array of information from internet resources and literature. For complex problems, the decompositional ability of LLMs is employed to construct associated questions. Further, incorporating multi-input/multi-output schemas and meta-hint techniques facilitates the LLM in autonomously creating suggestive prompts that steer the synthetic data generation process.

Building upon these insights and identified enhancement opportunities, we propose a revised framework for BT generation tasks, formalized in the following updated Equation~\ref{eq:btgen_refined}:

\begin{equation}
\label{eq:btgen_refined}
\begin{aligned}
    S_{0} & = f_{init}(root), \\
    S_{t+1} & = f_{MCTS}(S_{t}, f_{RAG}(S_{t})),
\end{aligned}
\end{equation}

where \( f_{init} \) signifies the initial setup function that creates the root of the tree, \( f_{MCTS} \) denotes the enhanced planning function integrating Monte Carlo Tree Search, and \( f_{RAG} \) represents the Retrieval-Augmented Generation mechanism.

Expanding on our analysis, we introduce the BT generation method and an agent-based framework for its implementation. As portrayed in Figure \ref{fig:btgen_mcts}, the BT generation approach adopts a strategy patterned after the Monte Carlo Tree Search (MCTS), elucidating the progression of states throughout the process of BT generation.

In this schema, each BT is depicted as a state $S_t$. Initiated with only a root node at $S_0$, the BT generation algorithm iteratively develops $S_t$ into intermediate BTs that function yet may not entirely align with predefined objectives. The ultimate aim is to converge upon a final state $S_{final}$, reflecting a BT proficient in executing the intended task accurately.

The transition between states is choreographed using an MCTS-like strategy, composed of four cardinal stages: \textit{Selection}, \textit{Expansion}, \textit{Validation}, and \textit{Iterative Refinement}, all tailored to accommodate the nuances of BT generation.

During the \textit{Selection} phase, LLM-based strategies are utilized, especially those involving Retrieval-Augmented Generation (RAG) as previously mentioned. The 'Ops Library' contains a suite of actions that facilitate the transition from parent nodes to child nodes, including various decomposition operations like sequential and parallel constructs. Simultaneously, the Knowledge Base retains a catalog of actionable leaf nodes applicable in simulation or real-world environments. The result of the \textit{Selection} phase encompasses custom operations and task-specific nodes, laying the groundwork for subsequent expansion prompts.

Following this, the \textit{Expansion} phase engages LLMs to use the generated prompts to extend the structure of BT. This leads to the development of multiple sub-nodes, delineating the succeeding state.

Next, the \textit{Validation} phase rigorously evaluates the newly formed BT $S_{t+1}$, determining the viability of this progression. Should the expansion adhere to the pre-set benchmarks, the state transformation proceeds; if not, feedback is harnessed to guide additional \textit{Selection} and \textit{Expansion} cycles—this recursive process embodies \textit{Iterative Refinement}.

The process culminates when further decomposition of leaf nodes becomes untenable, signaling the creation of the final BT, denoted as $S_{final}$. The performance of the BT generation methodology hinges on these methodical transformations, collectively ensuring the systematic assembly of a capable and efficient BT.

\section{Training}\label{sec:training}
LLMs exhibit substantial improvements when trained on extensive datasets rich in domain-specific language and terminologies. Such targeted training enhances their capabilities in specialized applications that necessitate robust natural language understanding, including interaction with computing systems through tasks like program synthesis, code completion, debugging, and documentation generation.

In our BT generation method, as we analyzed, require various LLM abilities which are the most advanced usage of LLM, such as planning, reasoning, tool manipulation abilities, etc. 
Some maybe meet our task requirements after being pretrained but it is various on the different LLM model. And Some abilities can not meet our requirements. So to make LLM model more useble, the training is necessary.  
The quality of data is paramount for effective training and ethical development of LLMs, a principle equally relevant to the context of BT generation. Typically, these models leverage diverse and publicly accessible open-source data during their initial training phase.

In this section, we will delineate the pretraining and Supervised Fine-Tuning (SFT) stages within the LLM training pipeline.

\subsection{Pretraining}
The pretraining of LLMs is conventionally initiated using a broad dataset. However, to refine and elevate the linguistic capabilities of an existing model, additional pretraining is employed, targeting domain-specific data that may have been underrepresented during the initial training corpus compilation.

This pretraining phase is designed to bolster the model's comprehension by introducing it to more focused data sets that encapsulate the idiosyncrasies and jargon intrinsic to the tasks for which the model is being primed. The construction of this domain-oriented data for our BT generation task has been elaborated in the preceding section.
Similar to the original pretraining, this phase engages unsupervised or self-supervised learning strategies but narrows down the data spectrum to align with specific domains. For example, Masked Language Modeling (MLM) techniques are reutilized, prompting the model to infer masked segments of text, thus honing its grasp on the subtleties of specialized lexicons.

Although a substantial portion of the model's parameters was established during primary pretraining, the additional pretraining introduces refinements through the assimilation of novel datasets. The quintessential goal remains consistent: to reduce predictive inaccuracies and enable the model to adjust its extensive knowledge base to resonate more precisely with targeted fields of application.
This augmented pretraining stage is particularly vital when the intended applications possess linguistic features that are markedly different from those prevalent in the original pretraining data. By ensuring that the model enters the fine-tuning phase already attuned to the nuances of the tasks, this step substantially augments the efficiency of Supervised Fine-Tuning (SFT) and the overall performance of the LLM within domain-specific contexts.

The execution of this pretraining segment presents significant challenges, rooted not only in the arduous task of acquiring pertinent BT-related data but also in the intricate details of the training process that are critical for optimal performance. While some foundational work \cite{ roziere2023codellama} has illuminated aspects of these complexities, extensive empirical practice remains indispensable. Despite its inherent difficulties, this preparatory phase is essential, setting the groundwork for effective task-specific LLM applications.
  
\subsection{Supervised Fine-Tuning (SFT)}

In scrutinizing the training dynamics of foundation models such as Code LLaMa \cite{roziere2023codellama}, the predominant emphasis has been placed on refining the underlying base model. This focus underscores a belief that foundational enhancements are instrumental for long-term performance gains, relegating changes in inference mechanisms to a secondary role. However, the examination of intricate inference methodologies constitutes a compelling frontier in the research landscape. Initiatives in this domain have introduced the integration of structural programming insights into training objectives. For instance, bespoke objectives dedicated to code deobfuscation have been formulated \cite{lachaux2021dobf}, along with the application of contrastive learning using semantically invariant transformations of code \cite{jain2020contrastive}, and the adoption of Abstract Syntax Trees to impart tree-aware positional encodings \cite{peng2021integrating}.

An expanding area of research contemplates the utility of program execution or unit tests for enhancing the evaluation and reinforcement of program correctness, particularly when the pool of solutions is limited \cite{li2022competition, chen2022codet, zhang2023pgtd, le2022coderl}. In parallel, other studies have woven unit tests into the tapestry of reinforcement learning, thereby magnifying the training signal \cite{le2022coderl, liu2023rltf}.

Fine-tuning represents the process by which a model's broad capabilities, honed during pretraining, are tailored to meet the specific demands of a target task or domain, leading to substantial gains in targeted application efficacy.
As its name implies, supervised fine-tuning leverages supervised learning techniques, relying on labeled datasets that map inputs to their respective target outputs. This correlation enables the model to adjust its parameters with precision, optimizing it for the task at hand.
For SFT in our BT generation task, the data corpus exclusively comprises task-specific input-output pairs. Data collection spans multiple formats, each designed to engage different required abilities. The aim of this fine-tuning phase is to train the LLM to interpret user prompts and generate appropriate outputs across various formats such as Code, XML, or constrained languages.

Recent advancements in fine-tuning techniques have yielded methodologies like P-tuning, LoRA, and QLoRA, each offering unique contributions to the customization of models for particular tasks. P-tuning\cite{liu2021ptuning} introduces prompt tuning as a method of soft prompt engineering, where continuous vectors (prompts) are optimized to guide the model's predictions without altering the model's parameters.
LoRA\cite{hu2021lora}, which stands for low-rank adaptation, modifies the attention mechanism by introducing low-rank matrices to adapt the model's behavior while maintaining most of the pre-trained parameters intact. 
QLoRA\cite{dettmers2023qlora} takes this concept further by quantizing the low-rank matrices, reducing memory footprint and computational overhead even more significantly.
In practice, LoRA and QLoRA are the methods that can easy to train and deploy, yet keep the good performance of the LLM model. 
In practice, training packages such as Deepspeed \footnote{https://github.com/microsoft/DeepSpeed} provide useful and easy-to-setup tools to support the achievement of SFT. These packages utilize optimizations to make the training process fast, reliable and distributed.

These fine-tuning approaches enhance model specificity, outperforming their corresponding generic pretrained variants on specialized tasks. While the computational resources required for fine-tuning are considerable, they are generally less intense than those needed for comprehensive pretraining, given the head start provided by an existing pretrained model foundation.

\section{Developing}\label{sec:develop}

After thorough training and extensive testing to confirm that the BTGen Model satisfies our stringent capability criteria, we are now poised to outline the proposed approaches for generating BTs. These strategies capitalize on the LLM's robust features for effective deployment and real-world application.

We propose a BT generation application method based on LLMs. This method requires carefully crafted prompts to ensure that the LLM's responses align with specified criteria. To tackle the issues of hallucination and out-of-domain knowledge, we employ an agent-based framework that incorporates elements of memory, planning, action, and multi-agent interactions. This framework is specifically designed to manage the complexities involved in creating BTs and harnesses the strengths of LLMs to simulate real-world scenarios that robots might encounter.

\subsection{Prompting}

LLMs have evolved as instrumental assets across a plethora of Natural Language Processing applications, markedly when guided through judiciously crafted prompts.\footnote{Within academic circles, the conception of a "prompt" is multifaceted. For our discussion, it is defined as a directive for a specific task, such as "Translate English to French," or a structured cue like "Let's dissect this problem step by step."} The art of prompt engineering—meticulous and strategic development of prompts—is paramount, as it involves a nuanced selection of language that steers the LLM toward producing relevant and articulate outputs. This aspect becomes even more critical within the realm of BT generation for robotics, where proficient prompting directly influences the creation of effective BTs.

There exists an array of methodologies for devising prompts, with notable approaches being Zero-shot Prompting, Few-shot Prompting, Chain-of-Thought Prompting, and Prompts Optimization.

\textbf{Zero-shot Prompting:} Zero-shot prompting introduces a novel task to the language model without furnishing any preceding examples. Within the domain of BT Generation, zero-shot prompts must be succinct yet comprehensive, capturing an explicit representation of the required behavior, which includes delineating the problem at hand, the operative regulations, and the envisioned final state. An exemplary zero-shot prompt may declare, \textit{Design a BT for an autonomous robot tasked with prioritizing kitchen cleanliness whilst optimizing energy consumption.} In this scenario, the model leverages its preexisting knowledge to compose a viable BT without reliance on extraneous exemplars.

\textbf{Few-shot Prompting:} Advancing from zero-shot techniques, few-shot prompting equips LLMs with a handful of instances. These precedents, proven effective in the domain of coding tasks \cite{roziere2023codellama}, embody the target output structure or schema prior to soliciting the generation of novel content. For BT Generation, providing the system with rudimentary examples of BTs or schematics of antecedent tasks can bolster its adeptness in fabricating precise and contextually suitable BTs. Moreover, few-shot prompting can alleviate uncertainties inherent in the initial prompt and construct a sturdier scaffolding for the model to replicate, culminating in outcomes that are more uniform and reliable.

\textbf{Chain-of-Thought related Prompting: }
Advancements in LLMs have seen the emergence of novel prompting techniques that enhance their reasoning capabilities. A pivotal development is the Chain-of-Thought (CoT) prompting method introduced by Wei et al.\cite{wei2022cot}. This technique encourages LLMs to articulate intermediate reasoning steps, which increases transparency and performance on complex tasks.
Extending CoT's linear approach, Yao et al. proposed the Tree-of-Thoughts (ToT) framework\cite{yao2023tot}, enabling LLMs to perform multi-trajectory decision-making, a boon for tasks requiring substantial exploration or intricate planning. The ToT's branching structure better reflects the nonlinear nature of such problem-solving scenarios.
Complementing this, Besta et al. introduced the Graph-of-Thoughts (GoT) strategy\cite{besta2023got}, which organizes knowledge into a graph where nodes symbolize discrete information units interconnected by edges representing their relationships. This graph-based method encapsulates human cognitive complexity, offering an approach for its emulation within LLMs.
To facilitate rapid decision-making, Ning et al. developed the Skeleton-of-Thought (SoT) technique\cite{ning2023sot}, priming the generative process with broad strokes before detailing specifics. This hierarchical technique not only accelerates the response time of LLMs but also yields more diverse and pertinent outcomes.
In coding, Li et al. introduced the Structured-CoT (SCoT) \cite{li2023scot} and the Chain-of-Code (CoC) \cite{li2023coc} methodologies. SCoT uses structured patterns inherent in source code to guide LLMs, enhancing code generation accuracy. CoC integrates natural language, pseudocode, and executable code, featuring an LMulator, a dual-mode execution system combining traditional interpreters with a language model-based emulator, to maintain program state continuity even when code is non-executable.
These Chain-of-Thought methods aim to enable LLMs to comprehend complex tasks not directly solvable from inputs and stored knowledge, and to decompose them into manageable sub-steps. They employ specialized designs prompt LLMs to think sequentially (CoT), from coarse-to-fine (SCoT), or from code-to-solution (CoC), simulating human-like problem-solving. Representing a shift towards more organized and sophisticated reasoning, these methods allow LLMs to address problems with increased cognitive fidelity.

\textbf{Prompts Optimization: }
The design of prompts significantly affects the quality of generated BTs. 
It is crucial to check how the prompts affect the quality of outputs.
Studies have found that the selection of demonstrations included in prompts significantly impacts accuracy across most tasks \cite{liu2021makes, agrawal2022context, xu2023k}.
Lu \cite{lu2021fantastically} thinks that altering the ordering of a fixed set of demonstrations can affect downstream accuracy. 
Prompts sensitive to demonstration permutation often exhibit lower accuracies \cite{chen2022relation}, making them less reliable, particularly in low-resource domains.
InstructEval \cite{ajith2023instructeval} is an in-context learning evaluation suite to conduct a thorough assessment of these techniques.  
Using the suite, InstructEval evaluates the relative performance of seven popular instruction selection methods over five metrics relevant to in-context learning.

\textbf{For the BT Generation Task.}
To produce functional and organized BTs, it is essential to create prompts that accurately reflect the structural intricacies of BTs, which include both composite nodes (sequences, selectors, parallels) and leaf nodes (actions or conditions). A well-designed prompt should clearly outline the objectives of the agent, the environmental factors, and any behavioral constraints while incorporating domain-specific terminology and constructs. This ensures that the resulting BTs are specific and adhere to the norms of automated planning and artificial intelligence. \textbf{Prompts Optimization} drawing inspiration from recent works such as InstructEval, aim to optimize our designed prompts by testing them on existing LLMs. evaluate the prompts using accuracy and sensitivity metrics. Accuracy measures whether BTs run without compilation and achieve the desired functions, while sensitivity uses the standard deviation of accuracies obtained with varying selection or permutation of the demonstrations. The average scores of accuracy and sensitivity serve as the final performance indicators for the designed prompts.

BTs organize the execution of action nodes similarly to how lines of code are executed in programming, highlighting control flow and data exchange among nodes. Their hierarchical nature suits the sequential processes found in CoT methods, which decompose goals into smaller tasks, map out possible actions, and assess them against current conditions. To leverage CoT techniques in generating BTs, we advocate two main ideas:
\textbf{Utilizing Control Structures:} Drawing from SCoT method, initially use a large language model to discern the control structure from an abstract task description. This forms the basis for creating a logical BT. By combining this structure with task context, the LLM can build a BT architecture that embodies a logical sequence of behaviors to achieve the robotic objectives.
\textbf{Aligning to Code Generation:} Aligning with the CoC approach, synthesize the BT by first framing a solution in code format using an LLM. This facilitates effective problem interpretation and leads to the translation of the coded solution into a BT. This technique leverages the parallel between coding logic and BT structures, ensuring the generated BTs accurately represent the required action sequences and controls for autonomous robot operation.

\subsection{Agent Technologies}

One significant problem with LLMs is their production of \textit{hallucinated} content—responses that appear coherent but are not factually accurate or relevant. This issue is particularly critical when LLMs are used to create BTs, which turn abstract commands into specific actions for robots, Unmanned Aerial Vehicles (UAVs), and other hardware systems. In such applications, the exactness and domain-specific knowledge are vital because any errors in transforming digital instructions into physical actions can cause malfunctions or inefficiency.

Agent technology leads the way in overcoming these challenges by being more than a collection of advanced methods—it's an integrated approach that improves the effectiveness of the entire system. Integrating LLMs within agent frameworks has enriched question-answering systems significantly, representing a crucial advance in artificial intelligence. To develop dynamic and adaptive autonomous agents that can operate and grow within intricate environments, it's important to design sophisticated architectures.

Recent strides in integrating LLMs with agent-centric frameworks provide compelling evidence of progress toward achieving artificial general intelligence across multiple domains. For instance, the \textbf{Ghost in the Minecraft (GITM)} project~\cite{zhu2023GITM} showcases how Generally Capable Agents (GCAs) can skillfully complete complex tasks like the \textit{ObtainDiamond} challenge in the open-world game Minecraft, leveraging text-based knowledge and memory without heavy reliance on GPU-intensive training approaches. Concurrently, the \textbf{Voyager} initiative~\cite{wang2023voyager} presents an LLM-inspired embodied agent exhibiting continuous learning and autonomous skill acquisition, surpassing existing benchmarks in tasks such as item collection, exploration, and mastering technology progression. Moreover, the \textbf{Plan4mc} program~\cite{yuan2023plan4mc} aims to overcome the inherent limitations of Reinforcement Learning (RL) in multitasking within expansive, procedural environments. By fusing skill-based reinforcement learning with strategic planning and intrinsic motivation, this approach fosters a diverse skill set and employs LLMs to craft a skill graph that streamlines task achievement. Additionally, \textbf{HuggingGPT}~\cite{shen2023hugginggpt} places LLMs at its core, orchestrating a consortium of specialized AI models emanating from communities like Hugging Face. This synergistic approach substantially enhances problem-solving capabilities across numerous domains and modalities using linguistic interfaces.

The contemporary trend in leveraging LLMs suggests a shift from merely solving tasks to engaging in elaborate planning and orchestration within complex scenarios. This new era of autonomous agent ability extends significantly beyond previous capabilities, supported by an advanced structure comprising four interrelated modules:
The Memory Module serves as the cornerstone for future decision-making by storing past interactions, akin to human memory. It's responsible for handling immediate sensory information, sustaining long-term knowledge, and includes reading, writing, and reflective processes that aid in the agent's evolution.
The Action Module integrates insights from other modules to generate concrete actions. It deals with intention setting, crafting action sequences, assessing possible actions, and reviewing outcomes, ensuring a coherent interaction between the agent and its environment through contemplation, execution, and evaluation phases.
The Planning Module equips the agent with strategic problem-solving skills. Mirroring human reasoning by breaking down complex issues into more manageable components, it offers static and dynamic planning techniques, the latter adjusting in response to environmental changes to preserve adaptability.
The Profile Module defines the agent's identity and role for specific applications, embodying various personas such as a coder or teacher. 

\subsubsection{Memory Module}

In the field of robotics, creating BTs requires specialized knowledge and the ability to adjust to changing environments. While LLMs are good at forming complex patterns, their training on fixed datasets limits them. To enhance LLMs in developing BTs, we've tried using carefully crafted prompts. However, this method often fails with new, unpredictable situations that were not part of the original data.

To address this issue, we can integrate a Retrieval-Augmented Generation (RAG) system, similar to how human memory works. This combines LLM with a database, letting the model access fresh and relevant information beyond what it learned initially. This "memory" stores past interactions, guiding the robot's future behaviors and decisions. Like our memory turns sensory input into lasting knowledge, this integrated system enables continual learning and adapting through \textit{read, write, and reflect} processes for novel experiences.

The field of BT generation stands on the cusp of transformation with emerging technologies such as Self Reflective Retrieval Augmented Generation (Self-RAG)\cite{asai2023selfrag}, Thought Propagation (TP)\cite{yu2023thoughtpropagation}, Sketch-based Automatic Code Generation (SkCoder)\cite{li2023skcoder}, and Automatic Multi-Step Reasoning and Tool-Use (ART)\cite{paranjape2023art}. Self-RAG enhances traditional LLM functionalities by integrating a self-evaluation tool, pivotal for refining dynamic, accurate, and context-aware BTs. In contrast, TP eliminates the need for extensive prompting by utilizing solutions from similar past issues, enriching the model's cognitive faculties for crafting more complex and applicable BTs. SkCoder employs an incremental approach familiar to human programmers, starting with basic code sketches and progressively enhancing them, aligning well with the iterative nature of developing BTs. Lastly, ART bolsters the model's problem-solving acumen by adding sequential reasoning and the use of external computational tools, enabling the construction of BTs that exceed the intrinsic limitations of the model itself.

The effects of RAG on LLMs are both beneficial and challenging.
On the one hand, incorporating RAG into LLMs enhances their performance by utilizing up-to-date external knowledge, improving context understanding and response precision. This can lead to better outcomes than models without access to such data integration \cite{lewis2020retrieval}.
However, using RAG introduces the risk of incorporating poor-quality data from external sources, which can lead to incorrect outputs \cite{kenton2019bert}. Over-reliance on one data source may cause overfitting and reduce the model's ability to generalize. Careful validation methods are essential to ensure that newly integrated information maintains the accuracy of LLMs.

\textbf{For the BT Generation Task.} It is critical to establish a comprehensive and accurate knowledge base. The effectiveness of a large language model relies heavily on accessing precise information, as inaccuracies can lead to irrelevant or incorrect output, particularly when designing BTs. A well-curated knowledge base enables the LLM to produce contextually relevant and factually sound BTs.
Incorporating the RAG approach into the generation pipeline is essential and involves two main aspects.
\textbf{Data Engineering:} The success of a RAG system depends on the quality of domain-specific data available. For creating BTs for specific scenarios, it's vital to have an extensive set of relevant facts organized effectively. If this data isn't prepared carefully, the resulting BTs may lack contextual relevance and fail to support effective planning.
\textbf{Retrieval Methods:} The retrieval process must ensure a close semantic match between the user's query and the knowledge base content. Semantic similarity metrics help find and extract information closely related to the query. The caliber of retrieved data significantly influences the LLM's performance in generating BTs. Poor retrieval could introduce irrelevant details, undermining the content's validity. Thus, sophisticated retrieval techniques are imperative to fetch the most pertinent information for accurate and useful BT generation.

\subsubsection{Action Module}

BTs offer a structured methodology for formulating high-level behaviors across various execution platforms, including robotic systems and Unmanned Aerial Vehicles. By representing behaviors as nodes within the BT, we abstract the intricacies of lower-level control mechanisms. In our approach, each node encapsulates a set of functionalities accessible via Application Programming Interfaces (APIs) or bespoke tools found in proprietary code libraries.
  
APIs are pivotal in software engineering, offering an abstraction that simplifies complex logic into more manageable, modular interfaces. This level of abstraction enables easier task delegation, diminishes interdependencies, and promotes modularity. LLMs to generate BTs involves prompting these models with specific API calls. In turn, the LLMs construct BTs that exhibit logical coherence and integrate seamlessly with extant systems or libraries. Through encapsulation, LLMs produce targeted snippets of code that conform to the unique specifications of the project at hand.

Several methods have been proposed for incorporating API usage into LLMs, thereby addressing related challenges. Search tools within frameworks such as ToolCoder~\cite{zhang2023toolcoder} and CRAFT~\cite{yuan2023craft} empower LLMs to identify and incorporate suitable APIs directly into generated code, ensuring precision and contextual appropriateness. These frameworks' toolsets support dynamic API retrieval and execution, which aids in crafting efficient code without necessitating model retraining. The LATM framework~\cite{cai2023latm} casts LLMs both as creators and consumers of tools, fostering continuous utility function generation and application. Such strategies enhance reuse and cost-efficiency; they enable advanced models to develop intricate tools and allow simpler models to use them effectively. Open-source projects like the Toolink framework~\cite{qian2023toolink} employ a chain-of-solving approach, wherein smaller models tackle specific tasks efficiently, rivaling proprietary alternatives at a fraction of the computational expense. This embodies a comprehensive strategy for the development and utilization of specialized toolsets and intelligent API integration, tailored for robust and flexible coding solutions in private domains.

\textbf{For the BT Generation Task,} its efficacy hinges upon the strategic configuration of selected nodes that are attuned to the requirements of the intended platform. These nodes constitute integral segments of code that direct the operations of robotics or artificial intelligence agents. It is imperative to meticulously establish these nodes to ensure proper functionality. The action module should rigorously consider the following domains:
\textbf{Node Libraries:} The construction of a comprehensive node library is a labor-intensive process that demands considerable reflection on the capabilities and constraints of the platform in question. Nodes must be tailored to align with the platform's functionalities, necessitating customization for seamless integration. The development of this library entails extensive strategic planning and an in-depth understanding of potential tasks the BT may be required to execute. Additionally, there is a need for flexibility to accommodate the creation of novel nodes in response to emergent behaviors.
\textbf{Node Selection:} After the assembly of the node library, the selection of the appropriate nodes becomes pivotal in sculpting the BTs to fulfill its designated purpose. The employment of Retrieval-Augmented Generation (RAG) can facilitate the identification of optimal nodes. This approach leverages insights gleaned from expansive databases to recommend prime nodes while generating new contextual information. Such methodologies enhance the relevance and quality of the nodes selected.
\textbf{Constrain the Results:} Ensuring that outcomes derived from Large Language Models are confined within the parameters of the pre-established node-set is a critical consideration. This constraint is vital to maintain systemic integrity and to guarantee the operability of BTs on their respective platforms. Integrating the BTs with the execution systems is essential for devising a coherent action plan. Furthermore, this forms the basis for subsequent phases of Validation — which ascertains the accurate functioning of the BTs — and Execution, whereby the AI or robotic entity purposefully engages with its environment.

\subsubsection{Planning Module}

The planning module is a critical element within the architecture of an autonomous agent, highlighting the substantial influence that LLMs exert in the realm of robotics, especially concerning the BT generation. LLMs are at the frontier of propelling robotic systems to interpret contexts and perform tasks autonomously, an attribute essential to genuine intelligence.

Forging a rational and executable strategy for problem-solving, whether it be through programming or natural language processing, is fundamental. Methods like the Chain-of-Thought process promote clarity in the reasoning trajectory, while RAG integrates specialized knowledge, increasing the precision of responses. The incorporation of these planning techniques into LLMs strengthens their ability to devise organized multi-step solutions.

Historically, automated planning draws upon classical frameworks such as the Planning Domain Definition Language (PDDL) \cite{silver2023pddl} and the Stanford Research Institute Problem Solver (STRIPS), which utilize abstract representations and action-precondition schemes to generate plans sequentially. However, these conventional approaches may struggle in dynamic contexts due to their reliance on fixed models.
Contrastingly, modern developments in Convolutional Neural Networks and Reinforcement Learning have ushered in alternative planning paradigms like Monte Carlo Tree Search (MCTS)\cite{swiechowski2023mcts}, renowned for its sophisticated search capabilities. Despite their efficacy in controlled scenarios, these methods can become less reliable amidst incomplete data and the inherent unpredictability of real-world environments. Here, LLMs with their pattern recognition expertise have the potential to enhance analytical and decision-making processes when amalgamated with planning algorithms.

In the context of code generation, LLMs' planning faculties aid in comprehending and carrying out intricate commands. Models that split the task into separate planning and execution stages, along with methodologies such as Planning-Guided Transformer Decoding (PG-TD)\cite{zhang2023pgtd}, direct token generation toward more refined outputs. For robot task planning, directed graphs employed by LLMs facilitate the BT generation designed for complex environments. Approaches like Language Agent Tree Search (LATS)\cite{zhou2023lats} and Reasoning via Planning (RAP)\cite{hao2023rap} use tree search methods informed by environmental input to improve decision-making. The neuro-symbolic approach of the LLM-DP model\cite{dagan2023llmdp} outperforms standard planners in executing physical tasks, with recent research demonstrating GPT-4's adeptness in planning across various fields, underscoring the importance of automated debugging and summarization of Chain-of-Thought in this sphere.

\begin{figure*}[htbp]
    \label{fig:agent}
    \centering
    \includegraphics[width=0.8\textwidth]{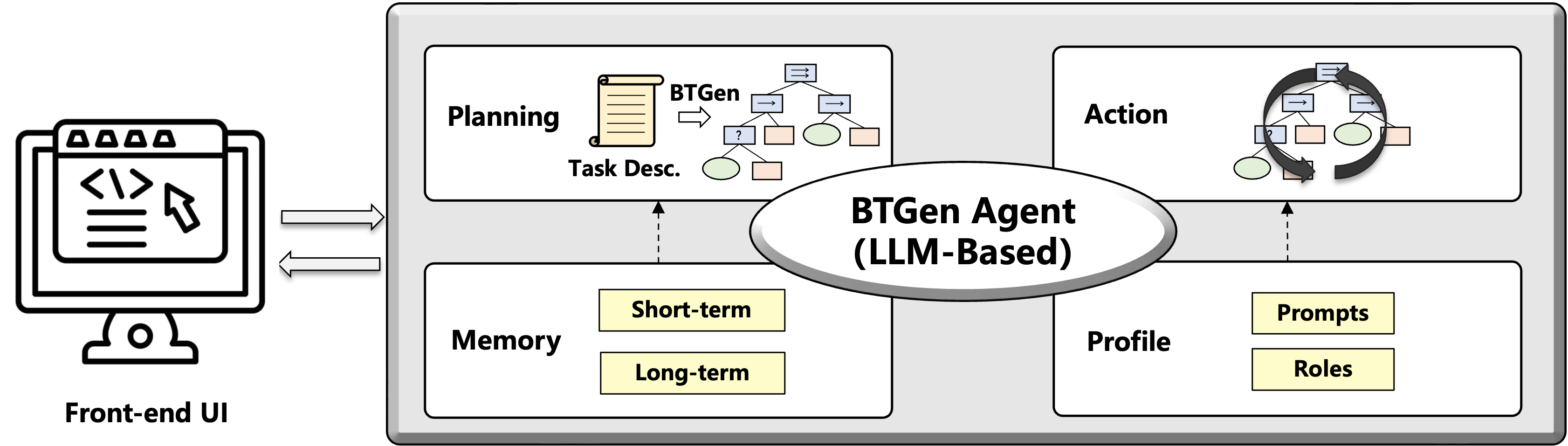}
    \caption{Illustration of the BTGen Agent framework.}
\end{figure*}

\textbf{For the BT Generation Task.} Embedding LLM-based planning into BT generation offers notable advantages. These trees systematically arrange actions, allowing agents to handle diverse situations competently. 
By integrating LLMs endowed with advanced planning capabilities, BTs can evolve dynamically, embodying authentic decision-making structures. 
Sophisticated planning strategies within LLMs, such as those utilized by LATS and RAP, confer strategic insight upon BTs, enabling them to adapt to environmental shifts and make forward-looking decisions. This engenders a system capable of complex, human-like reasoning and planning, thereby enhancing its robustness and flexibility. The merger of LLM-enabled planning and BT generation heralds a transformative era for autonomous systems management, preparing them to navigate intricate and volatile conditions with agility paralleling human cognition.
The intertwined progress of LLMs and planning methodologies promises to redefine the autonomy of future systems, signifying a stride towards an AI that reflects the nuance and complexity of human thought processes.

\subsubsection{Profile Module}

The versatility of LLMs lies in their capacity to adopt varied personas under the direction of specific prompts. Within an agent-based framework, LLMs can function across multiple modules, each with a unique application. The Profile module is pivotal in managing diverse roles during computational processes.

Recent research highlights the effective use of LLMs as team members in software development. \cite{dong2023selfcollaboration} introduces a self-collaborative system where several ChatGPT instances serve as experts in specific fields such as analysis and testing, mimicking human teamwork to simplify processes and improve code quality autonomously.
Similarly, \cite{qian2023communicative} presents ChatDev, a virtual platform following the waterfall model, where communicative agents facilitate a seamless workflow through dialogue in all stages of software production. This method is cost-effective and expedites development while proactively addressing errors.
Furthermore, the DyLAN framework proposed by \cite{liu2023dylan} enhances LLM agent collaboration with a flexible interaction paradigm that forms optimal teams for tasks, especially in complex areas like logic and code creation. By assessing agent significance without supervision, DyLAN consistently outperforms single-agent models.
Together, these advancements suggest a shift towards harnessing LLM agents' collective intelligence in automated software development, combining diverse cognitive skills with increased efficiency.

\textbf{For the BT Generation Task.} the LLM adopts distinct personas including a planner, who segments the task into smaller actionable subtasks, and a validator, who ascertains the accuracy of outputs and provides feedback for iterative enhancements, among others. The Profile module employs a multi-agent perspective to organize these varying roles, managing the assortment of prompts required to navigate the task's pipeline effectively.

\subsubsection{Refinement: beside Agent}

Refinement describes the process of incrementally enhancing computational outputs through cycles of evaluation and modification. Echoing human solution-refining practices, it employs a trial-and-error approach akin to programmers' iterative debugging and optimization of code.

This concept, though not new, has been pivotal in the recent progress seen in LLM advancement. For instance, the Self-Refine model \cite{madaan2023selfrefine} illustrates the capability of LLMs to engage in self-improvement cycles, autonomously refining their outputs. Such mechanisms are integral to cutting-edge models, including GPT-4, highlighting the importance of self-evaluation. Additionally, Reflexion \cite{shinn2023reflexion} introduces linguistic feedback loops into language learning, further refining the decision-making processes in language agents. These feedback mechanisms have made significant strides in code generation tasks, suggesting advances in AI systems that adapt with limited training data. In robotic task planning, the ISR-LLM (Iterative and Self-Refining LLM) model \cite{zhou2023isrllm} integrates a verification phase in its iterative cycle, verifying the robustness and feasibility of proposed action plans. Similarly, the STOP (Self-Training with Optimistic Planning) framework \cite{zelikman2023stop} promotes recursive self-improvement, illuminating the prospects for self-evolving AI systems.

\textbf{For the BT Generation Task.} It is unrealistic to expect perfect LLM outputs from the onset. Instead, iterative refinement serves as a strategy to progressively improve LLM-generated BTs, with an emphasis on executability and functional coherence. The initial output from the LLM acts as a baseline, subject to consecutive rounds of testing, evaluation, and adjustment. Drawing inspiration from successful LLM frameworks, such as ISR-LLM, we endeavor to facilitate the initial generation and continuous refinement of generated BTs. The central component of the refinement is the validator, which functions similarly to a code executor, acting as a world model to execute and offer feedback on the generated BTs. The iterative cycle includes automated simulations or manual assessments, tailored to specific domain requirements, involving detailed adjustments to the BTs and corresponding API calls to meet performance benchmarks.

\subsection{Our BTGen Agent}

Building upon the established agent framework, we have devised a BT generation framework intended to create execution BTs, named \textbf{BTGen Agent}. 
It comprises four main modules: memory module, action module, planning module, and profile module. 
In addition to these modules, it also incorporates a refinement mechanism to ensure usability.

\textbf{Memory Module:}
The memory module serves as the linchpin of our agent-based execution BT generation framework. Its purpose is to accumulate and utilize essential knowledge efficiently to construct sophisticated execution BTs. The design accommodates dynamic and rapid access to domain-specific knowledge required for tree generation. Comprising several components, it includes Short-Term Memory (STM), which retains transient contextual expertise from expert input, ongoing process cues, and initial LLM reasoning outcomes; Long-Term Memory (LTM), a durable store that aggregates and safeguards the agent's accumulated experiences and strategies, compatible with scalable vector databases for complex data sets and structured databases; Prompt Optimization and Retrieval Enhancement, which hone the LLM outputs to align with predefined leaf node ontologies and operational parameters of the agent, enriching its retrieval capabilities using long-term memory.

\textbf{Action Module:}
The action module, essential in verifying the execution BT's ability to achieve the anticipated task performance, translates abstract task outlines into executable actions. It refines task descriptions and provides action nodes adaptable to diverse software and hardware platforms, designed for both scalability and practicality. Its three vital elements are the Action Node Ontology for categorizing and decoding various actions, Action Node Linking that binds leaf nodes to executable segments or APIs, and the Action Node Repository, which houses updatable resources for new tasks and platforms. Through these components, the module underscores the execution of the effectiveness of BTs and supports seamless functioning across different settings.

\textbf{Planning Module:}
The planning module amalgamates LLMs' advanced planning abilities with classic planning algorithms, facilitating the transformation of user-supplied task descriptions to execution BTs. It leverages outputs from the Memory Module and employs MCTS and CoT algorithms to build precise and effective trees. By incorporating RL, the module iteratively optimizes tree generation based on feedback from the agent's interactions with its environment, consequently enhancing plan quality and efficiency. Furthermore, the alignment of tree generation with code generation tasks improves complexity and accuracy, enabling coherent high-level reasoning and meticulous execution of complex commands by the platform.

\textbf{Profile Module:}
The profile module enables the generation of execution BTs across varied roles by configuring LLMs using distinct prompt templates. This Configuration Module designs role-specific prompts to excavate and amplify LLMs' performance in knowledge extraction, task decomposition, tool selection, and verification driving. Real-time prompt adjustments permit the model to adapt to different roles seamlessly, optimizing performance to suit operational scenarios within complex task environments. This methodology has significantly improved the LLMs' capacity in key areas, ensuring their applicability to the dynamic requisites of intricate tasks.

\textbf{Refinement Mechanism:} 
The refinement in our BTGen Agent is not an independent module, but rather a mechanism that ensures the correct execution of the BT generation task by incorporating feedback from the simulation. The refinement occurs at multiple levels. During the generation of the BTs, this refinement is fast, inexpensive, but not entirely accurate. It provides affordable yet effective feedback to guide the generation process. Additionally, the execution of the BTs is necessary for the final output. During the validation stage, it evaluates whether the final BTs can achieve the target goal. The evaluation results may lead to discarding the generated behavior and initiating a new generation. This multi-level refinement is key to generating useful and correct BTs.

\subsection{Application}
In this section, we detail the framework and methodologies employed to develop BT generation pipelines within production settings. Our emphasis is on the dual aspects of the application framework enabling such deployments and the evaluation of performance metrics crucial for a seamless, responsive user experience.
The proposed application framework is an amalgam of back-end services complemented by a front-end user interface. These components synergistically facilitate the creation, administration, and application of BTs in robotic systems.

\textbf{Back-end Services} is the cornerstone of our approach, encompassing a comprehensive suite of APIs that provide robust support for the front-end interaction. These APIs are designed to handle a spectrum of operations that include the BT generation, iterative refinement of these structures, integration of executable action nodes or ancillary tools, and the employment of rigorous validation mechanisms to ensure structural and functional integrity. The Back-end architecture is conceived to function as the computational backbone of the system—processing complex requests and returning results with minimal latency, thus fostering a dynamic, resilient environment conducive to sophisticated manipulations of BTs.

\textbf{Front-end User Interface (UI)} is architected to offer an intuitive, streamlined user engagement with BTs. Through the UI, users can access a diverse toolkit enabling tasks like the execution of BTs, verification of their architectural coherence and logical soundness, and graphical representation to elucidate and share concepts. The UI aids in real-time editing, simulates behavior execution, and provides analytics regarding performance. Collectively, these features empower users to efficiently iterate over the development cycle, optimizing the BTs for deployment in robots.

\section{Verification and Validation}\label{sec:vv}

V\&V plays an important role in training and developing BTGen model and they ensure the functions correctly while meeting the specified requirements.
On the one hand, \textit{verification} focuses on checking whether the BTGen model and the generated BTs are built correctly according to the original design.
On the other hand, \textit{validation} focuses on checking whether the BTGen model can output the desired outcomes and the generated BTs could effectively perform the desired tasks.
According to Figure \ref{overview}, V\&V should cover all stages of training and developing the BTGen model.
We summarize them from two aspects: V\&V for abilities of BTGen models and V\&V for the performance of BTs.

\begin{table*}[htbp]
  \centering
  \renewcommand\tabcolsep{2pt}
  \footnotesize
  \caption{Benchmarks and metrics of verifying and validating abilities of BTGen model}
    \begin{tabularx}{\textwidth}{X|X|X|X}
    \hline\hline
    \textbf{Categories} & \textbf{Abilities} & \textbf{Metrics} & \textbf{Benchmarks} \\
    \hline
    \multirow{4}[0]{*}{\textbf{Natural-Language-Related}} 
    & Natural language understanding
    & \multirow{4}[0]{*}{\tabincell{l}{accuracy, F1-scores, perplexity, \\ROUGE-L}}
    & GLUE\cite{wang2018glue}, SuperGLUE\cite{wang2019superglue}, CLUE\cite{xu2020clue} \\
    \cmidrule{2-2}\cmidrule{4-4}
    & Natural language generation
    &
    & LMSYS-Chat-1M\cite{zheng2023lmsys}, MT-bench\cite{zheng2023judging}, LongBench\cite{bai2023longbench} \\
    \cmidrule{2-2}\cmidrule{4-4}
    & In-context learning
    &
    & \multirow{2}[0]{*}{\tabincell{l}{COIG\cite{zhang2023chinese}, 
    Flan\cite{weifinetuned}, Flan 2022\cite{longpre2023flan},\\
    OPT-IML\cite{iyer2022opt}, SuperNI\cite{wang2022super}
    }}\\
    \cmidrule{2-2}
    & Instruction following
    &
    \\
    \hline
    \multirow{2}[0]{*}{\textbf{Reasoning-Related}} 
    & Commonsense reasoning
    & \multirow{3}[0]{*}{\tabincell{l}{accuracy, F1-scores}}
    & CommonsenseQA\cite{talmor2018commonsenseqa}, PIQA\cite{bisk2020piqa}, Pep-3k\cite{bisk2020piqa}, Social IQA\cite{sap2019socialiqa}, HellaSWAG\cite{zellers2019hellaswag}\\
    \cmidrule{2-2}\cmidrule{4-4}
    & Logical reasoning
    & 
    & Alpaca-CoT\cite{si2023empirical}, LSAT\cite{wang2022lsat}, MultiNLI\cite{williams2017broad}, LogicNLI\cite{tian2021diagnosing}, ConTRoL\cite{liu2021natural}, ReClor\cite{yu2020reclor}, LogiQA\cite{liu2020logiqa}, LogiQA 2.0\cite{liu2023logiqa} \\
    \cmidrule{2-2}\cmidrule{4-4}
    & Planning
    & 
    & PlanBench\cite{valmeekam2023planbench} \\
    \hline
    \textbf{Tool-Related} 
    & Retrieval
    & precision, recall, F1-scores
    & \multirow{2}[0]{*}{\tabincell{l}{
    API-Bank\cite{li2023api}, APIBench\cite{patil2023gorilla}, \\
    ToolBench\cite{xu2023tool}, ToolLLM\cite{qin2023toolllm}}
    }\\
    \cmidrule{2-3}
    & Tool manipulation
    & accuracy, F1-scores, ROUGE-L, pass rate, success rate\\
    \hline
    \textbf{Code-Related} 
    & Code generation
    & \multirow{3}[0]{*}{\tabincell{l}{pass@k, perplexity, ROUGE-L}}
    & HumanEval\cite{chen2021evaluating}, MBPP\cite{austin2021program},
    APPS\cite{hendrycks2021measuring}, MultiPL-E\cite{cassano2023multipl}, EvalPlus\cite{liu2023your},
    HumanEval-X\cite{zheng2023codegeex}\\
    \cmidrule{2-2}\cmidrule{4-4}
    & Code explanation
    & 
    & CodeXGLUE \cite{lu2021codexglue}\\
    \cmidrule{2-2}\cmidrule{4-4}
    & Code translation
    & 
    & XLCoST \cite{zhu2022xlcost}, HumanEvalPack \cite{muennighoff2023octopack}\\
    \hline\hline
    \end{tabularx}%
    \label{metrics_benchmarks}
\end{table*}%

\subsection{V\&V for Abilities of BTGen Models}

These abilities have been introduced and required for the selection of foundation LLMs in Section \ref{sec:foundation_models}.
However, we still check these abilities in verifying and validating the BTGen model due to two reasons.
First, these abilities are crucial for the BT generation task and they may be degraded during the steps of training and developing the BTGen model.
We need to ensure that the BTGen model always has these abilities.
Second, we also introduce the BT related benchmarks and metrics to assist verifying and validating these abilities.
Thus, to generate the high quality BTs, we need to check these abilities of the BTGen model in V\&V.
We summarize the metrics and benchmarks may be used in Table \ref{metrics_benchmarks}.

Most natural-language-related tasks and reasoning-related tasks could be divided into two classes: classification tasks and generation tasks.
In terms of the classification tasks, such as sentiment analysis, text classification, natural language inference, the inputs consist of a single sentence or sentence pairs, while the output labels denote the specific classes or indicate whether the two sentences are related. 
Since their outputs are class labels, the \textit{accuracy} and \textit{F1-scores} are always adopted as the metrics.
In terms of the generation tasks, common tasks include translation, summarization and multi-turn dialogues.
These tasks are challenging primarily due to diversity and flexibility.
\textit{ROUGE-L} (Recall-Oriented Understudy for Gisting Evaluation-Longest Common Subsequence) is an evaluation metric commonly used in natural language process and text summarization tasks. 
It is designed to measure the similarity between a generated summary and a reference summary.
Moreover, \textit{perplexity} is also adopted in text generation to assess the effectiveness of LLMs. 
It is a way to measure how surprised or uncertain the model is when trying to predict the next token in a sequence. 
Mathematically, \textit{perplexity} is defined as the inverse probability of the test set, normalized by the number of words (or tokens) in the test set, which can be calculated using Eq \ref{perplexity}.
\begin{equation}
    \label{perplexity}
    \small
    Perplexity= \sqrt[N]{\frac{1}{P(w_1, w_2, \cdots, w_N)} }
\end{equation}
In this equation, $N$ is the number of tokens in the test set and $P(w_1,w_2,\cdots, w_N)$ is the probability assigned by the language model to the test set.

The current V\&V methods on tool-related abilities primarily focus on following three aspects:
\textbf{Assessing retrieval} evaluates the retrieval ability involves assessing how well the system can find and present the most relevant information in response to a query or user need. 
Common metrics used to evaluate the retrieval ability are \textit{precision}, \textit{recall}, and \textit{F1-scores}.
\textbf{Assessing feasibility} assesses whether the model can successfully execute the given tools by comprehending them.
Commonly used evaluation metrics in this dimension include the execution \textit{pass rate} and tool operation \textit{success rate}.
\textbf{Assessing performance} evaluates how well it performs once it has been established that the model can achieve the task. 
It examines the correctness of the final answer by calculating the \textit{accuracy}, \textit{F1-scores}, and \textit{ROUGE-L} depending on specific tasks.

\begin{figure}[htbp]
    \centering
    \includegraphics[width=3in]{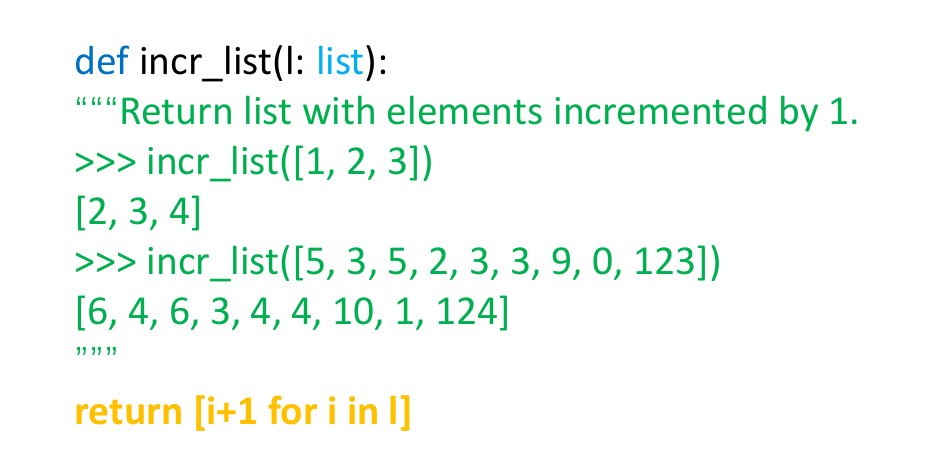}
    \caption{An example for the HumanEval benchmark. 
    The prompt provided to the model is shown in \textcolor{green}{green}, and a successful model-generated completion is shown in \textcolor{Dandelion}{orange}.
    }
    \label{humaneval}
\end{figure}

In terms of the code-related abilities, \textit{pass@k} is the most important metric to evaluate the correctness of generated program codes.
\cite{kulal2019spoc} evaluates functional correctness using the widely adopted metric, \textit{pass@k}, where $k$ code samples are generated per problem, a problem is considered solved if any sample passes the unit tests, and the total fraction of problems solved is reported.
In each problem, the LLMs generate $n\geq k$ samples per task, counts the number of correct samples $c\leq n$ which pass unit tests, and calculates the unbiased estimator in Eq \ref{pass_at_k}.
HumanEval\cite{chen2021evaluating} has been widely adopted in assessing the ability of code generation by many research works and leaderboards\footnote{https://www.datalearner.com/ai-models/llm-coding-evaluation,https://huggingface.co/spaces/bigcode/bigcode-models-leaderboard,https://evalplus.github.io/leaderboard.html}. 
Each problem in HumanEval, as shown in Figure \ref{humaneval}, provides a prompt with descriptions of the function to be generated, function signature, and example test cases in the form of assertions. 
The LLM needs to complete a function given the prompt such that it can pass all provided test cases, thus measuring the performance by functional correctness.

\begin{equation}
    \label{pass_at_k}
    \small    
     pass@k=\mathbb{E}_{problems} \left [ 1-\frac{C_{n-c}^{k}}{C_{n}^{k}}\right ] \\
\end{equation}

\begin{figure}[htbp]
    \centering
    \includegraphics[width=3.5in]{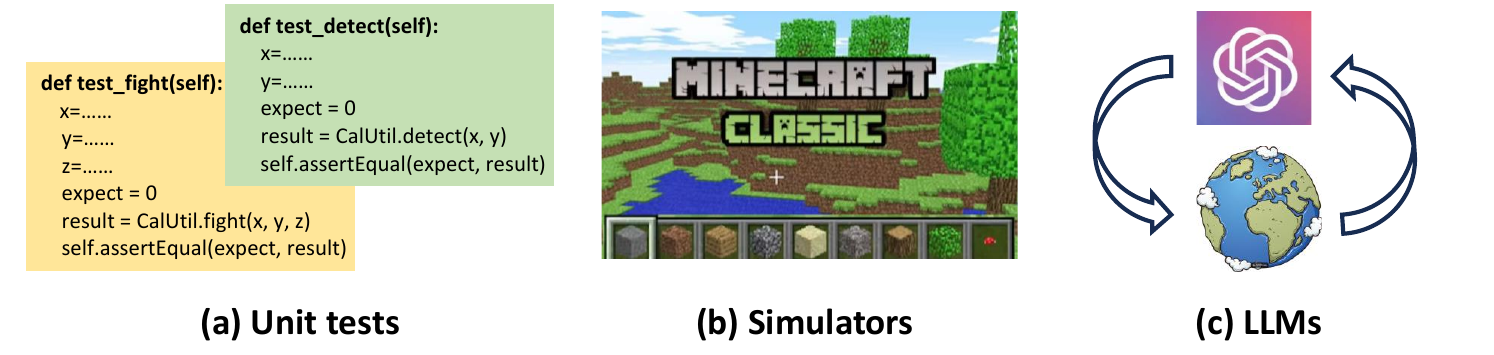}
    \caption{V\&V of BTGen model could be classified into three categories based on unit tests, simulators, and LLMs.}
    \label{BT_vv}
\end{figure}

\subsection{V\&V for the Performance of BTs}

Second, V\&V checks the performance of BT to improve the credibility of generated BTs and further adjusts the generated BTs.
As shown in Figure \ref{BT_vv}, existing V\&V methods could be summarized into three categories based on unit tests, simulators, and LLMs.
We will introduce each of them as follows.

\subsubsection{Unit-Tests-Based V\&V}

Unit tests are essential practices in software development that involves testing individual units or components of a software system in isolation, including function testing, class testing, module testing, integration testing, and code coverage testing.
They target various input scenarios of the function and check whether the function returns the expected outputs correctly.

As for testing BTs, these unit tests could test the BT nodes, tree execution, and logic integration.
First, each node in a BT can be considered a unit, and thus unit tests can be written to verify that each node performs its intended function correctly.
For example, the action nodes that perform specific tasks can be tested to ensure they handle edge cases and errors properly.
Second, unit tests can be constructed to test the correct execution of the BTs itself. 
These tests can verify that the tree traversal is happening correctly, that nodes are being executed in the proper order, and that the correct branches are taken based on different conditions.
Third, when a BT is part of a larger application, unit tests can be utilized to ensure that the BT interacts with other application components properly. 
This might involve mocking external dependencies and testing the behavior of the tree under various simulated conditions.
Common test tools, such as \textit{JUnit}\footnote{https://junit.org/junit5/}, \textit{GoogleTest}\footnote{https://google.github.io/googletest/} and manually crafted a set of unit tests checks whether the corresponding functions in generated BTs are effective.

\subsubsection{Simulator-Based V\&V}

Simulator, as a computational tool, encompasses the emulation of real-world processes or systems by employing mathematical formulas, algorithms, or computer-generated representations to imitate their behaviors or characteristics \cite{gao2023large}.
The significance of simulation spans various domains, serving as a valuable tool for understanding, analyzing, and predicting intricate phenomena that might be impractical or impossible to observe directly in real life.
Simulators aims to simulate the transition of states based on specified rules and object properties, which has been widely adopted in many tasks \cite{wang2023voyager,gong2023mindagent, shridhar2020alfworld, yang2023appagent}.
Voyager \cite{wang2023voyager} is a lifelong learning agent in Minecraft simulator\footnote{https://www.minecraft.net/} that continuously explores the world, acquires diverse skills, and makes novel discoveries without human intervention.
It is able to utilize the learned skill library in a new Minecraft world to solve novel tasks from scratch, while other techniques struggle to generalize.
Moreover, Zhu \cite{zhu2023GITM} also aims to create Generally Capable Agents (GCAs) in Minecraft.
MindAgent \cite{gong2023mindagent} establishes a new gaming scenario and related benchmark based on a multi-agent virtual kitchen environment, CuisineWorld. 
It adopts a minimal text-based game format and supports various planning task structures and difficulties, making it an ideal test bed for the emergent multi-agent planning (scheduling and coordination) capacity.
In addition, ALFWorld \cite{shridhar2020alfworld} is a simulator that enables agents to learn abstract, text-based policies in TextWorld \cite{cote2019textworld} and then execute goals from the ALFRED benchmark \cite{shridhar2020alfred} in a rich visual environment.
Specifically, TextWorld is an engine for interactive text-based games and ALFRED is a large scale dataset for vision-language instruction following in embodied environments.
Similar thoughts exist in \cite{puig2018virtualhome} where Puig introduces the VirtualHome simulator that allows researchers to create a large activity video dataset with rich ground-truth by using programs to drive an agent in a synthetic world.

By using simulators, we can simulate various scenarios and situations to observe the behavior and decision-making of agents under different conditions. 
However, building a high-quality simulator is not an easy task and requires significant time and resources. 
First, it is essential to gather data from the real world to create a precise simulation setting. This step might encompass laborious tasks like amassing data, refining it, and adding descriptive labels. 
Second, the creation and execution of the simulation's physics, behavior, and perception models is critical to guarantee the fidelity and dependability of the simulated context. 
Executing these steps demands much specialized knowledge.

\subsubsection{LLM-Based V\&V}

LLMs are a breaking achievement in the field of machine learning, demonstrating vast potential in tasks related to natural language understanding and text creation. 
Utilizing their impressive skills, LLMs are regarded as world models to represent the physical world, capable of simulating changes in the world's state in rseponse to various actions.
They offer potential improvements to simulations through facilitating more sophisticated and lifelike portrayals of decision-making, communication, and adaptability in simulated scenarios.

Test cases generation from LLMs is a practical scenario of LLM-based V\&V to aid the software development.
Code LLaMA \cite{roziere2023codellama} constructs the self-instruction dataset in the following: 
(1) Generate interview-style programming questions by prompting LLaMA-2 and de-duplicate the set of questions. 
(2) Generate unit tests and ten Python solutions by prompting Code LLaMA. 
(3) Run the unit tests on the ten solutions and add the first solution that passes the tests to the self-instruct dataset.
Similar thoughts could be seen in \cite{chen2022codet, huang2023codecot}.
They use the generated test cases to validate the effectiveness of generated codes.
Other works have attempted to directly replace simulators with LLMs.
\cite{de2023llmr} employs GPT-4 as an inspector for compilation and run-time error.
If errors are found, the inspector provides suggestions for corrections. 
The process iterates until either the code passes inspection or a maximum number of inspections is reached. 
\cite{hao2023rap} uses LLMs to understand the current environment state, comprehend the impact of actions on the environment state, and output the next environment state, thereby simulating state transitions. 

LLMs can serve as a new paradigm for simulation with human-level intelligence.
Integrating LLMs into simulation holds the potential to enrich the fidelity and complexity of simulations, potentially yielding deeper insights into system behaviors and emergent phenomena for the following reasons \cite{gao2023large}:
(1) LLMs can adaptively react and perform tasks based on the environment without predefined explicit instructions. 
(2) LLMs have strong intelligence to respond realistically and even actively take actions with self-oriented planning and scheduling. 
However, effectiveness and accuracy of LLM-based V\&V still require further research and investigation.

\begin{figure}[htbp]
    \centering
    \includegraphics[width=3.5in]{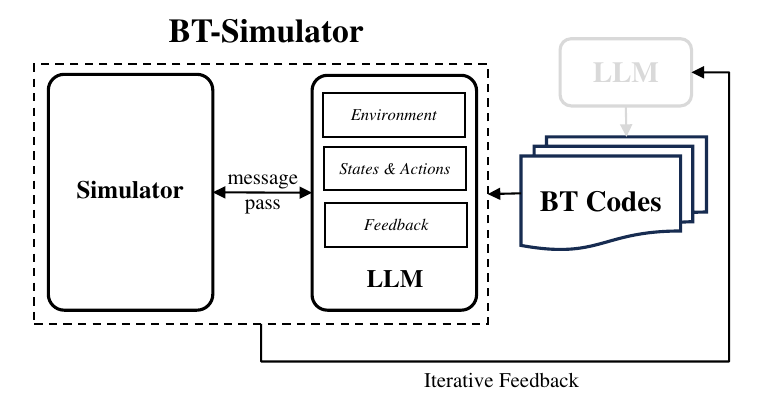}
    \caption{The framework of BT simulator in the loop of training and developing BTGen model.}
    \label{bt_evaluator}
\end{figure}

\subsection{Our Method for V\&V}

As shown in the above analysis, our method for V\&V could be divided into two steps.

First, we plan to introduce a benchmark for evaluating basic abilities of BTGen model in BT generation.
It evaluates these natural-language-related, reasoning-related, tool-related, and code-related abilities.
Specifically, to measure the performance of BTGen model in natural language understanding, natural language generation, in-context learning, and instruction following, we select CLUE \cite{xu2020clue}, MT-Bench \cite{zheng2023judging}, COIG \cite{zhang2023chinese}, and Flan 2022 \cite{longpre2023flan} as benchmarks to evaluate the BTGen model in terms of \textit{accuracy}, \textit{F1-scores}, and \textit{ROUGE-L}.
To measure the performance of BTGen model in commonsense reasoning, logical reasoning, and planning, we select CommonsenseQA \cite{talmor2018commonsenseqa}, Alpaca-CoT \cite{si2023empirical}, PlanBench \cite{valmeekam2023planbench} to evaluate the BTGen model in terms of the same metrics.
ToolBench\cite{xu2023tool} and APIBench\cite{patil2023gorilla} are selected to measure the BTGen model in the tool-related tasks i terms of the \textit{accuracy}, \textit{F1-scores}, \textit{ROUGE-L}, \textit{pass rate}, and \textit{success rate}.
At last, we measure the code-related abilities, including code generation, code explanation, and code translation of BTGen model on HumanEval\cite{chen2021evaluating} and CodeXGLUE \cite{lu2021codexglue} in terms of \textit{pass@k}.

Second, we measure the performance of BTs in simulation environments.
Li \cite{li2023coc} proposes an LMulator as a portmanteau of LLMs and code emulator.
If the generated code is successfully executed, the program state is updated and the execution continues. 
Otherwise, the language model instead is used to simulate the execution. 
The program state is subsequently updated by the outputs of LLMs and the execution continues. 
Inspired by LMulator, we plan to propose a novel preliminary idea of BT simulator that interweaves the existing simulator with LLM in a feedback loop, dealing with both semantic and algorithmic reasoning.
The detailed process could be shown in Figure \ref{bt_evaluator}.
On the one hand, existing simulators can provide a basic framework for simulating.
On the other hand, LLMs with world knowledge could assist the simulator achieving more generalization in different unseen scenarios and reduce labor overhead in designing simulators.
Specifically, the intermediate message information can be passed between the simulator and LLM to provide more valuable information.
LLM can exert its advantages in the following three aspects: setting environments, simulating states transition and actions generation, and designing feedback functions. 
Moreover, the BT simulator exists in the loop of BT generation where the outputs of it are regarded as the feedback to iteratively update the LLM to generate higher quality of BT codes.

\section{Open Questions}\label{sec:open_questions}

\textbf{Data Question.}
The success of LLMs like ChatGPT is largely dependent on the quality of data they are trained with. Acquiring high-caliber, representative datasets is a significant concern in this domain. Synthetic data generation has its merits, including scalability and customization; however, it also carries intrinsic limitations. Notably, when LLMs create data for a less advanced target model, the output may seem satisfactory but can lead to "hallucination issues"—instances where the data contains baseless components due to the training limitations of the source LLM.

Moreover, publicly sourced datasets are prone to contain 'dirty data'—information that is erroneous or extraneous—which can propagate bias or inaccuracies in the models. This highlights the crucial need for extensive data-cleaning processes and the establishment of benchmarks for dataset quality. Striking a balance between the volume and integrity of data is intricate, necessitating precise attention in order to advance the creation of more precise and reliable LLMs. Future research should concentrate on devising sophisticated methodologies for automated data validation and refinement, thereby ensuring models are trained on a robust and credible dataset.

\textbf{Training Question.}
In addition to data selection, the training strategies implemented for LLMs present a nuanced challenge. The complexities of pretraining and subsequent supervised fine-tuning have significant ramifications for an LLM's performance, which requires a deep understanding of the interplay between model scale and generalization capabilities. The appropriate configuration of hyperparameters and optimization techniques during fine-tuning is critical for task-specific adaptation, while avoiding pitfalls such as overfitting or catastrophic forgetting.

Further exploration is warranted to identify optimal practices for transfer learning in the LLM landscape, potentially employing meta-learning paradigms to promote quick acclimatization to new tasks or domains. There is also potential for innovation in refining LLM architectures, possibly through introducing modular elements or variations in attention mechanisms, which could improve the efficiency and effectiveness of training protocols. An in-depth examination of these training approaches is imperative to clarify the factors contributing to the superior efficacy of certain LLMs relative to others under comparable conditions.

\textbf{Planning Question.}
In the context of reasoning processes within LLMs, planning serves as a fundamental component for BT generation tasks. The input consists of the task objective, while the output manifest as structured executable actions in the form of a BT that robotic systems can enact. Presently, despite the demonstrated promise of LLMs in planning functions, there is a conspicuous disparity between these nascent skills and the necessities of practical applications.

To narrow this divide, research should focus on augmenting LLMs' understanding of sequential and causal relationships pertinent to tasks. The assimilation of external knowledge repositories and simulation platforms may provide ancillary support for grounding LLM-formulated plans in actuality. Moreover, it is essential to evolve from static planning models to accommodate dynamic settings, where real-time decisions are essential. Investigations should pivot toward composite models that blend LLMs' generative strengths with deterministic systems to manage the unpredictability of real-world circumstances effectively. Enhancing LLMs with contextual awareness and sensitivity may lead to innovative planning methodologies that reflect the intricacies of robotic operations more accurately.

\textbf{Validation Question.} 
A primary concern within our research is the verification of correctness in the outputs produced. Efficient validation of generated BTs represents a complex balancing act. We suggest that LLMs could streamline this process and offer novel insights compared to traditional simulators.

There is a pressing requirement for the development of stringent testing regimes capable of consistently assessing the functionality and safety of BTs. Such protocols should not merely test the logical consistency of sequences but also their responsiveness to environmental fluctuations or goal modifications. Additionally, establishing metrics for evaluating the interpretability and explicability of the trees is paramount, enabling end-users to trust and utilize them competently. Incorporating feedback mechanisms from actual deployments into the validation process may yield critical information for continuous enhancement.

Looking ahead, it is vital to explore the integration of simulation technologies with LLMs to craft virtual testing grounds where BTs can undergo rigorous assessment across diverse scenarios. This approach would facilitate comprehensive and economical validation prior to implementation in tangible robots, thus reducing risk and bolstering the dependability of autonomous actions. In essence, the goal is to cultivate a synergistic dynamic between LLM-guided design and empirical evaluations, potentially revolutionizing the standards of robotic system validation.

\section{Conclusions}\label{sec:conclusion}
In conclusion, this paper has provided a comprehensive exploration of the nascent field of LLMs utilization for BT Generation. By harnessing the power of LLMs, we have showcased the potential to significantly enhance the adaptability, generality, and interpretability of BTs. Our investigation into the deployment of LLMs within BT generation reveals that it is possible to streamline this task.

The proposed phase-step prompt design and the use of self-instructive datasets represent pioneering steps towards more user-friendly and intuitive methodologies for BT generation. These approaches could potentially alleviate common issues such as hallucination, data bias, out-of-domain limitations, and lack of explainability and transparency in current models.

Furthermore, our work emphasizes the importance of V\&V driven pipeline designed to consistently yield high-quality BTs. This pipeline serves to ensure that all stages of training and development meet stringent criteria for functionality and compliance with specified requirements.

It is imperative to acknowledge that while this field is still in its early developmental stages, the foundational work outlined in this paper sets forth a robust framework for future research. The strategies and insights presented here aim to catalyze progress, inviting further investigation and refinement of LLM applications in BT generation. As we advance, the intersection of AI automation and BT systems stands to gain significant innovations, ultimately benefiting a wide array of developers and researchers across diverse domains.

{\small
\bibliographystyle{unsrt}
\bibliography{literature}
}

\end{document}